\documentclass[10pt,twocolumn,letterpaper]{article}
\usepackage{cvpr}
\usepackage{times}
\usepackage{epsfig}
\usepackage{graphicx}
\usepackage{amsmath}
\usepackage{amssymb}
\usepackage{subfig}
\usepackage{caption}
\usepackage{float}
\usepackage{multirow}
\usepackage{xcolor}
\usepackage{makecell}
\usepackage{capt-of}
\usepackage[percent]{overpic}
\usepackage{threeparttable,booktabs}
\usepackage{etoolbox}
\usepackage[labeled,resetlabels]{multibib}
\appto\TPTnoteSettings{\footnotesize}
\usepackage{xfp}
\newcommand\SupplementaryMaterials{%
  \xdef\presupfigures{\arabic{figure}}% save the current figure number
  \xdef\presupsections{\arabic{section}}% save the current section number
  \renewcommand\thefigure{S\fpeval{\arabic{figure}-\presupfigures}}
  \renewcommand\thesection{S\fpeval{\arabic{section}-\presupsections}}
  \renewcommand{\thetable}{S\arabic{table}}
  \renewcommand{\theequation}{S\arabic{equation}}
}

\newcites{S}{Supplementary References}

\usepackage[pagebackref=true,breaklinks=true,colorlinks=true,bookmarks=false]{hyperref}

\definecolor{applegreen}{rgb}{0.0, 0.5, 0.0}
\definecolor{bblue}{HTML}{0063C8}

\cvprfinalcopy % *** Uncomment this line for the final submission

\def\cvprPaperID{****} % *** Enter the CVPR Paper ID here

\ifcvprfinal\fi
\begin{document}

%%%%%%%%% TITLE
\title{Refining activation downsampling with SoftPool}

\author{Alexandros Stergiou$^{1}$ \qquad Ronald Poppe$^{1}$ \qquad Grigorios Kalliatakis$^{2}$ \\
\hfill\linebreak[3]\\
\begin{tabular}{c c c c}
\normalsize{$^{1}$Utrecht University} & \normalsize{$^{2}$University of Warwick}\tabularnewline
\normalsize{Utrecht} & \normalsize{Coventry} \\
\normalsize{The Netherlands} & \normalsize{United Kingdom} \\
\tt\footnotesize{\{a.g.stergiou,r.w.poppe\}@uu.nl} & \tt\footnotesize{grigorios.kalliatakis@warwick.ac.uk}
\end{tabular}
}

\maketitle
%\thispagestyle{empty}

%%%%%%%%% ABSTRACT
\begin{abstract}
   Convolutional Neural Networks (CNNs) use pooling to decrease the size of activation maps. This process is crucial to increase the receptive fields and to reduce computational requirements of subsequent convolutions. An important feature of the pooling operation is the minimization of information loss, with respect to the initial activation maps, without a significant impact on the computation and memory overhead. To meet these requirements, we propose SoftPool: a fast and efficient method for exponentially weighted activation downsampling. Through experiments across a range of architectures and pooling methods, we demonstrate that SoftPool can retain more information in the reduced activation maps. This refined downsampling leads to improvements in a CNN's classification accuracy. Experiments with pooling layer substitutions on ImageNet1K show an increase in accuracy over both original architectures and other pooling methods. We also test SoftPool on video datasets for action recognition. Again, through the direct replacement of pooling layers, we observe consistent performance improvements while computational loads and memory requirements remain limited\footnote{Code is available at: \url{https://git.io/JL5zL}}.
\end{abstract}

%%%%%%%%% BODY TEXT

\begin{figure}[ht]
\centering
\includegraphics[width=.75\linewidth]{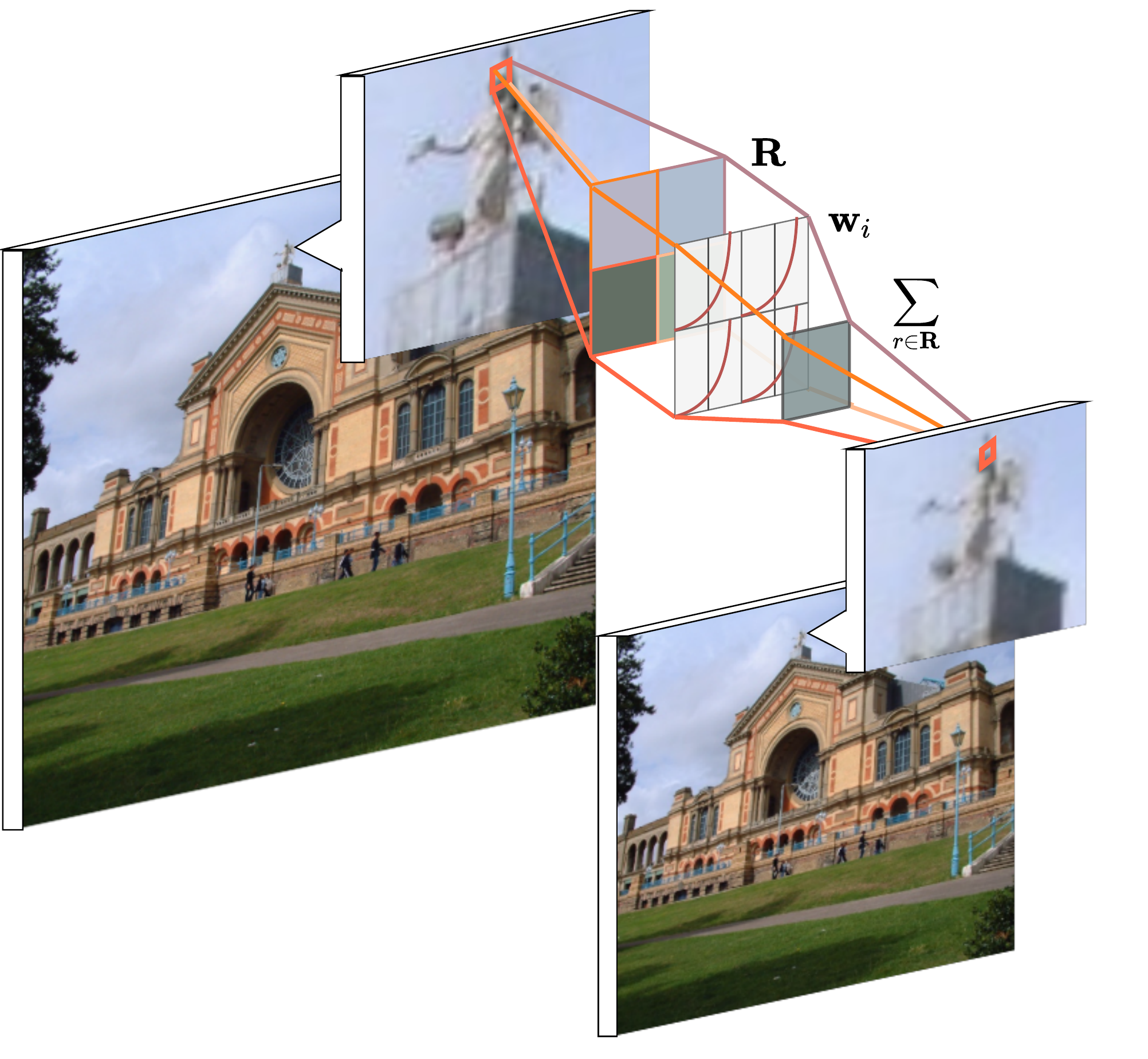}
\caption{\textbf{SoftPool illustration}. The original image is sub-sampled with a $2 \! \times \! 2$ ($k\!\!=\!\!2$) kernel. The output is based on the exponentially weighted sum of the original pixels within the kernel region. This can improve the representation of high-contrast regions, present around object edges or specific feature activations.}
\label{fig:softpool_concept}
%\vspace{-1em}
\end{figure}

\section{Introduction}
\label{sec:intro}

% Importance of pooling in CNNs
Pooling layers are essential in convolutional neural networks (CNNs) to decrease the size of activation maps. They reduce the computational requirements of the network while also achieving spatial invariance, and increase the receptive field of subsequent convolutions \cite{chen2017dual,real2019regularized,xie2017aggregated}.

% Challenge of pooling
%Pooling operators strive to limit the loss of information in the activation maps produced by sub-sampling while, at the same time, we favor lower computation and memory overheads. 
A range of pooling methods has been proposed, each with different properties (see Section~\ref{sec:related}). Most architectures use maximum or average pooling, both of which are fast and memory-efficient but leave room for improvement in terms of retaining important information in the activation map.

\begin{figure*}[ht]
\centering
\includegraphics[width=\textwidth]{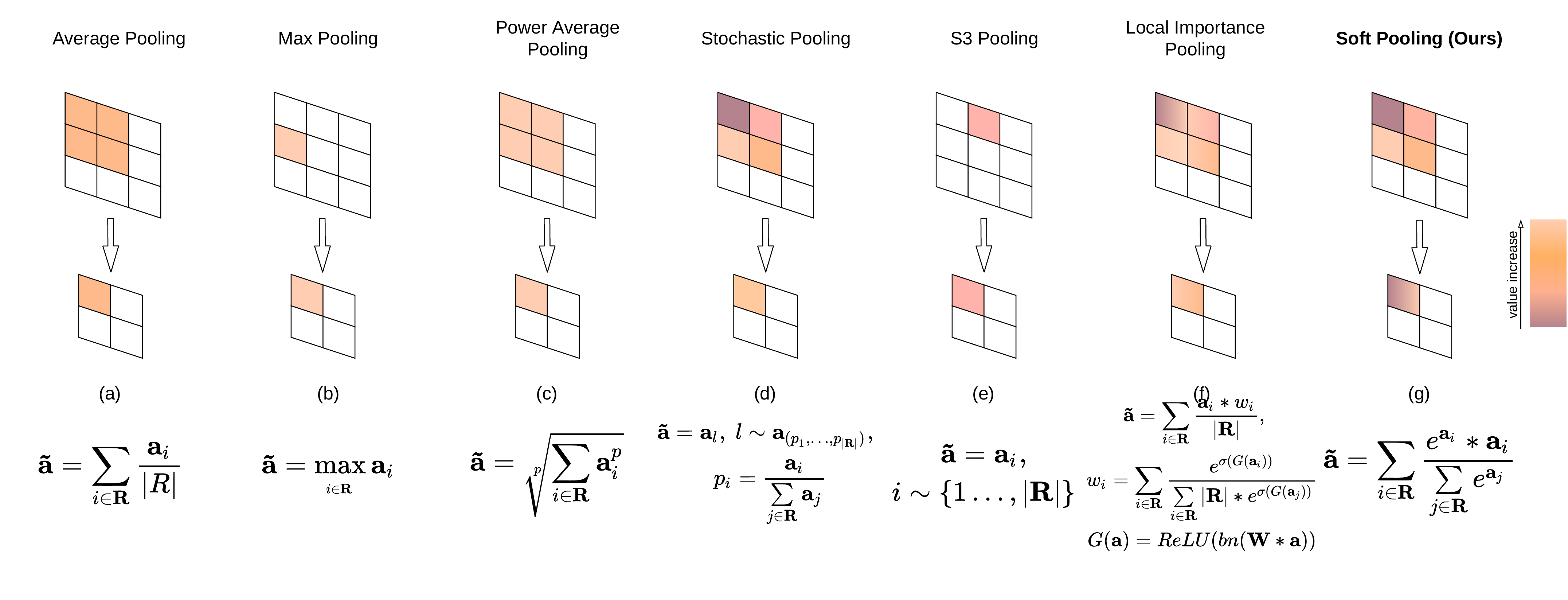}
\caption{\textbf{Pooling variants}. $\mathbf{R}$ is the set of pixel values in the kernel neighborhood. (a,b) \textbf{Average and maximum pooling} are based on averaging and maximum activation selection in a kernel. (c) \textbf{Power Average pooling} ($L_{p}$)~\cite{estrach2014signal,gulcehre2014learned} is proportional to average pooling raised to a power ($p$). The output is equal to max-pool for $p \rightarrow \infty$, and sum pooling when $p = 1$. (d) \textbf{Stochastic pooling}~\cite{zeiler2013stochastic} outputs a randomly selected activation from the kernel region. (e) \textbf{Stochastic Spatial Sampling} (S3Pool)~\cite{zhai2017s3pool} samples random horizontal and vertical regions given a specified stride. (f) \textbf{Local Importance Pooling} (LIP)~\cite{gao2019lip} uses a trainable sub-net $G$ to enhance specific features. (g) \textbf{SoftPool} (ours) exponentially weighs the effect of activations using a softmax kernel.}
\label{fig:softpool_dist}
\vspace{-1em}
\end{figure*}

% SoftPool and contributions
We introduce \textit{SoftPool}, a kernel-based pooling method that uses the softmax weighted sum of activations. We demonstrate that SoftPool largely preserves descriptive activation features, while remaining computationally and memory-efficient. Owing to better feature preservation, models that include SoftPool consistently show improved classification performance compared to their original implementations. We make the following contributions:
\begin{itemize}
    \item We introduce SoftPool: a novel pooling method based on softmax normalization that can be used to downsample 2D (image) and 3D (video) activation maps.
    \item We demonstrate how SoftPool outperforms other pooling methods in preserving the original features, measured using image similarity.
    \item Experimental results on image and video classification tasks show consistent improvement when replacing the original pooling layers by SoftPool.
\end{itemize}

% Paper outline
The remainder of the paper is structured as follows. We first discuss related work on feature pooling. We then detail SoftPool (Section~\ref{sec:methodology}) and evaluate it in terms of feature loss and image and video classification performance over multiple pooling methods and architectures (Section~\ref{sec:experiments}).

\section{Related Work}
\label{sec:related}

%In this section, we first discuss the importance of pooling for hand-coded features and then explain how pooling became a key component of CNNs. Finally, we present an extensive overview of pooling variants used in CNNs.

% Pooling in hand-coded methods
\textbf{Pooling for hand-crafted features}. Downsampling is a widely used technique in hand-coded feature extraction methods. In Bag-of-Words (BoW) \cite{csurka2004visual}, images were viewed as collections of local patches, pooled and encoded as vectors \cite{wang2010locality}. Combinations with Spatial Pyramid Matching \cite{lazebnik2006beyond} aimed to preserve spatial information. Later works considered the selection of the maximum SIFT features in a spatial region \cite{yang2009linear}.
Pooling has primarily been linked to the use of max-pooling, because of the feature robustness of biological max-like cortex signals \cite{serre2005object}. Boureau \textit{et al.} \cite{boureau2010theoretical} studied maximum and average pooling in terms of their robustness and usability, and found max-pooling to produce representative results in low feature activation settings.

% Pooling with learned features
\textbf{Pooling in CNNs}. Pooling has also been adapted to learned feature approaches, as seen in early works in CNNs \cite{lecun1998gradient}. 
The main benefit of pooling has traditionally been the creation of condensed feature representations which reduce the computational requirements and enable the creation of larger and deeper architectures \cite{simonyan2014very,szegedy2015going}.

% Pooling variants (plug-and-play)
Recent efforts have focused on preserving relevant features during downsampling. An overview of a number of popular pooling methods appears in Figure~\ref{fig:softpool_dist}. Initial approaches include stochastic pooling \cite{zeiler2013stochastic}, which uses a probabilistic weighted sampling of activations within a kernel region. Mixed pooling based on maximum and average pooling has been used either probabilistically \cite{yu2014mixed} or through a combination of portions from each method \cite{lee2016generalizing}. Based on the combination of averaging and maximization, Power Average ($L_{p}$) pooling \cite{estrach2014signal,gulcehre2014learned} utilizes a learned parameter $p$ to determine the relative importance of both methods. When $p=1$, the local sum is used, while $p\rightarrow \infty$ corresponds to max-pooling. More recent approaches have considered grid-sampling methods. In S3Pool \cite{zhai2017s3pool}, the downsampled outputs stem from randomly sampling the rows and columns of the original feature map grid. Methods that depend on learned weights include Detail Preserving Pooling (DPP, \cite{saeedan2018detail}) that uses average pooling while enhancing activations with above-average values. Local Importance Pooling (LIP, \cite{gao2019lip}) utilizes learned weights as a sub-network attention-based mechanism. Other learned pooling approaches such as Ordinal Pooling \cite{deliege2019ordinal}, which order kernel pixels in discerningly and assigning them trainable weights. More recently, Zhao and Snoek \cite{zhao2021liftpool} proposed a pooling technique named LiftPool based on the use of four different learnable sub-bands of the input. The produced output is composed by a mixture of the discovered sub-bands.

\begin{figure*}[ht]
\includegraphics[width=\textwidth]{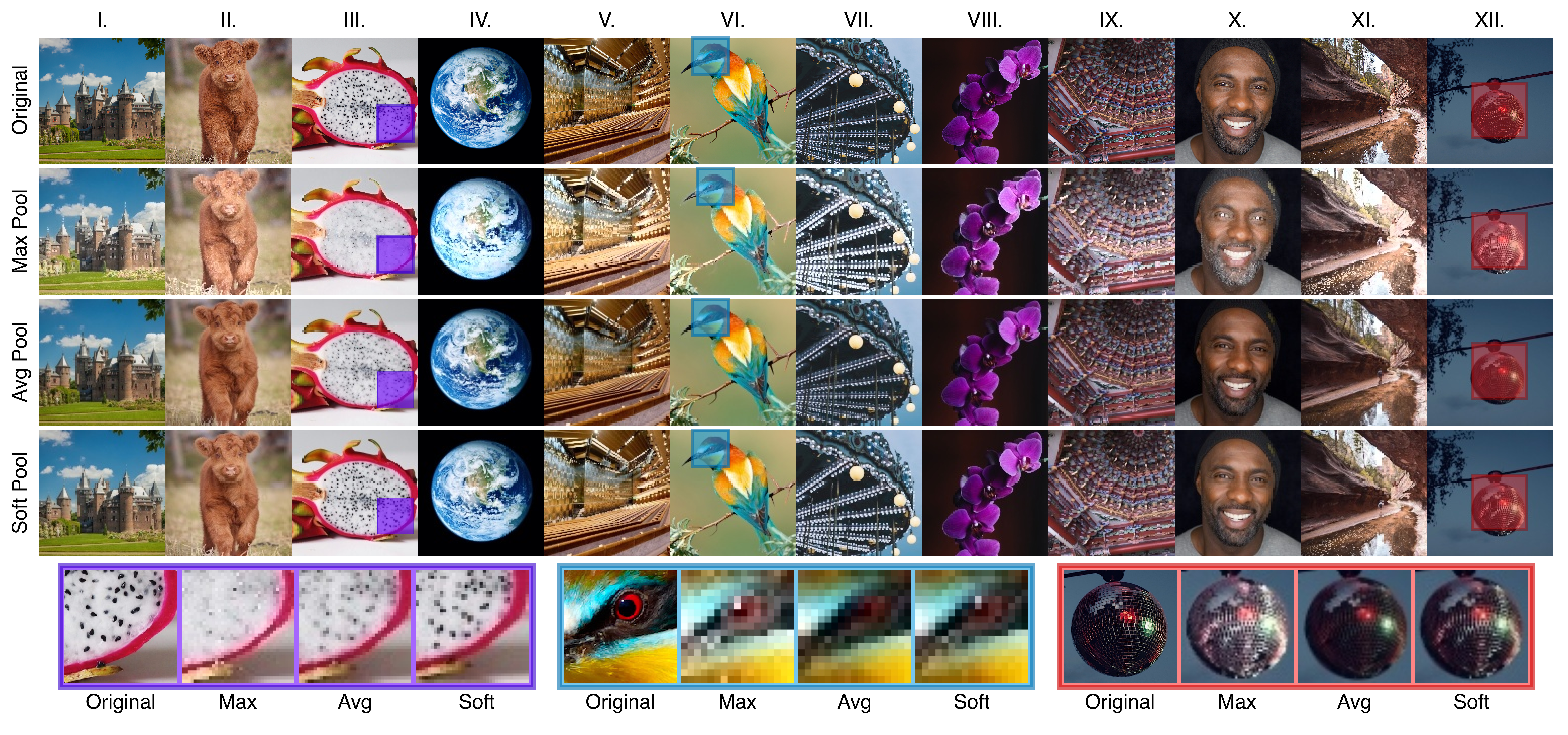}
\caption{\textbf{Examples of maximum, average and soft pooling}. Images are $1200 \times 1200$ pixels with the $3\times$ pooled equivalents reduced to 12.5\% of the original size. Image selection was based on overall contrast (I, III, V, VIII, XII), color (II, IV, VI, IX), detail (III, V, X, XI, XII) and texture (II, VII). Images on the bottom row are zoomed-in regions. A high resolution version of the regions appears alongside a detailed discussion in \textsection \textcolor{red}{1} of the Supplementary Material. %In most cases, max-pooling significantly distorts image features by increasing high-valued pixels (seen at the dragon fruit and disco ball examples II \& XII). Average pooling can decrease the neighborhood of pixels (as with the bird example VI). SoftPool preserves both the global structure, as well as informative details.
}
\label{fig:softpool_examples}
\vspace{-.9em}
\end{figure*}

% Shortfalls of current approaches
Most of the aforementioned methods rely on different combinations of maximum and average pooling. Instead of combining existing methods, our work is based on a softmax weighting approach to preserve the basic properties of the input while amplifying feature activations of greater intensity. SoftPool does not require trainable parameters, thus is independent to the training data used. Moreover, it is significantly more computational and memory efficient compared to learned approaches. In contrast to max-pooling, our approach is differentiable. Gradients are obtained for each input during backpropagation, which improves neural connectivity during training. Through the weighted softmax, pooled regions are also less susceptible to vanishing local kernel activations, a common issue with average pooling. We demonstrate the effects of SoftPool in Figure~\ref{fig:softpool_examples}, where the zoomed-in regions show that features are not completely lost as with hard-max selection, or suppressed by the overall region through averaging.

% Main methodology
\section{SoftPool Downsampling}
\label{sec:methodology}

We start by formally introducing the forward flow of information in SoftPool and the gradient calculation during backpropagation. We consider a local region ($\mathbf{R}$) in an activation map ($\mathbf{a}$) with dimension $C \times H \times W$ with $C$ the number of channels, $H$ the height and $W$ the width of the activation map. For simplicity of notation, we omit the channel dimension and assume that $\mathbf{R}$ is the set of indices corresponding to the activations in the 2D spatial region under consideration. For a pooling filter of size $k \times k$, we consider $|\mathbf{R}| = k^2$ activations. The output of the pooling operation is $\tilde{\mathbf{a}}_R$ and the corresponding gradients are denoted with $\nabla\mathbf{\tilde{a}}_i$.

\subsection{Exponential maximum kernels}
\label{sec:methodology::softmax}

% Building softmax kernels
SoftPool is influenced by the cortex neural simulations of Riesenhuber and Poggio  \cite{riesenhuber1999hierarchical} as well as the early pooling experiments with hand-coded features of Boureau \textit{et al.} \cite{boureau2010theoretical}. The proposed method is based on the natural exponent ($e$) which ensures that large activation values will have greater effect on the output. The operation is differentiable, which implies that all activations within the local kernel neighborhood will be assigned a proportional gradient, of at least a minimum value, during backpropagation. This is in contrast to pooling methods that employ hard-max or average pooling. SoftPool utilizes the smooth maximum approximation of the activations within kernel region $\mathbf{R}$. Each activation $\mathbf{a}_i$ with index $i$ is applied a weight $\mathbf{w}_{i}$ that is calculated as the ratio of the natural exponent of that activation with respect to the sum of the natural exponents of all activations within neighborhood $\textbf{R}$:

\begin{equation}
\label{eq:weight}
    \mathbf{w}_i=\frac{e^{\mathbf{a}_i}}{\sum\limits_{j \in \mathbf{R}}e^{\mathbf{a}_j}}
\end{equation}

% Non-linear transform vs linear
The weights are used as non-linear transforms in conjunction with the value of the corresponding activation. Higher activations become more dominant than lower-valued ones. Because most pooling operations are performed in high-dimensional feature spaces, highlighting the activations with greater effect is a more balanced approach than simply selecting the average or maximum. In the latter case, discarding the majority of the activations presents the risk of losing important information. Conversely, an equal contribution of activations in average pooling can correspond to local intensity reductions by considering the overall regional feature intensity equally.

% Summation of weights and pixels
The output value of the SoftPool operation is produced through a standard summation of all weighted activations within the kernel neighborhood $\mathbf{R}$:

\begin{equation}
\tilde{\mathbf{a}} =\sum\limits_{i \in \mathbf{R}}\mathbf{w}_{i}*\mathbf{a}_{i}
\end{equation}

\begin{figure}[ht]
\centering
\includegraphics[width=\linewidth]{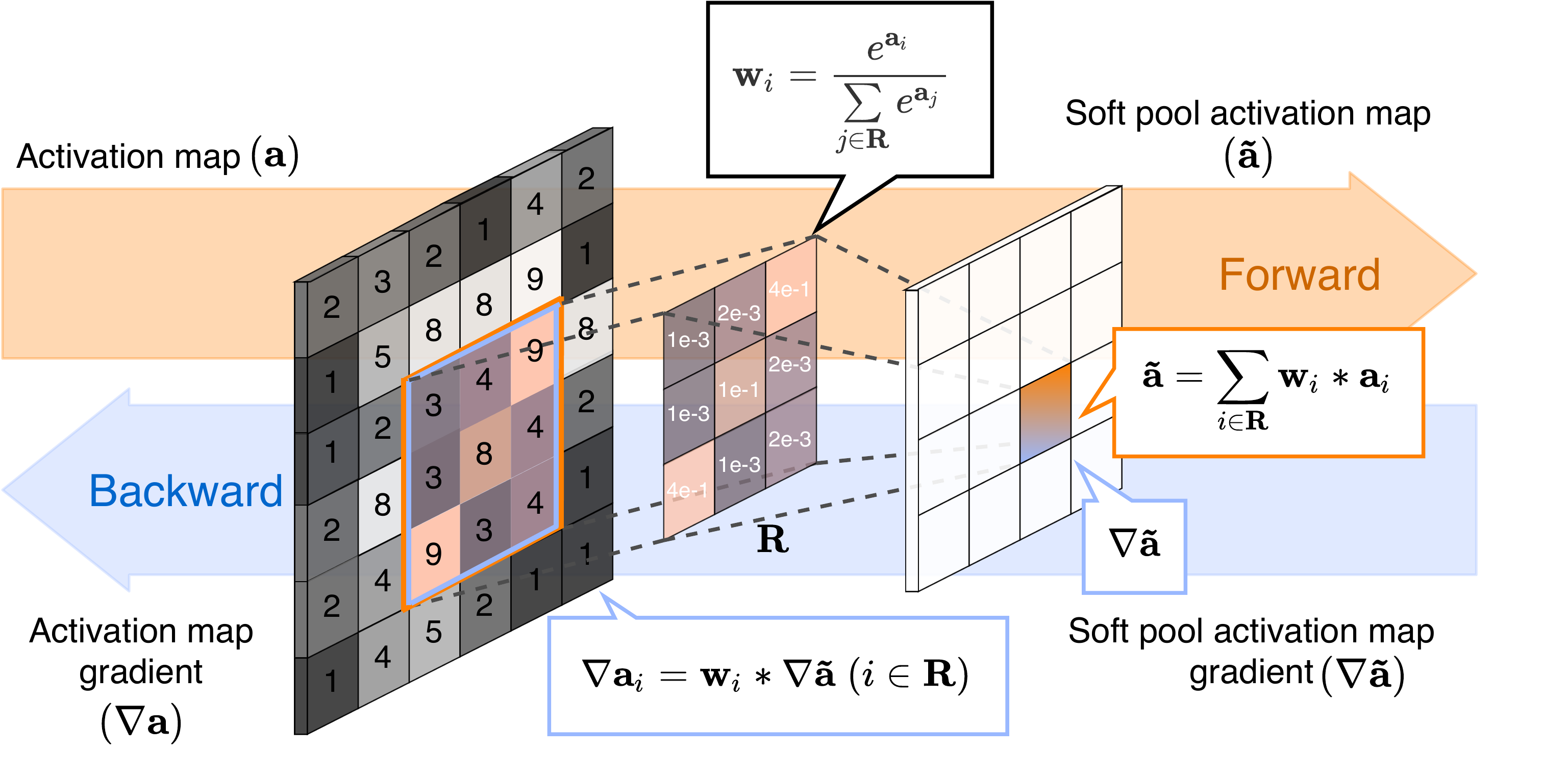}
\caption{\textbf{SoftPool calculation}. In forward operation, in \textcolor{orange}{orange}, the kernel uses the exponential softmax value of each activation as weight and calculates the weighted sum for region $\mathbf{R}$. These weights are also used for the gradients ($\nabla \tilde{\mathbf{a}}_{i}$), in \textcolor{bblue}{blue}. Activation gradients are proportional to the calculated softmax weights.}
\label{fig:softpool}
\vspace{-.5em}
\end{figure}

% Normalization through softmax
In comparison to other max- and average-based pooling approaches, using the softmax of regions produces normalized results with a probability distribution proportional to the values of each activation with respect to the neighboring activations for the kernel region. This is in direct contrast to popular maximum activation value selection or averaging all activations over the kernel region, where the output activations are not regularized. A full forward and backward information flow is shown in Figure~\ref{fig:softpool}.

\subsection{Gradient calculation}
\label{sec:methodology::gradients}

% Proportional gradients
During the update phase in training, gradients of all network parameters are updated based on the error derivatives calculated at the proceeding layer. This creates a chain of updates, when backpropagating throughout the entire network architecture. In SoftPool, gradient updates are proportional to the weights calculated during the forward pass.

% Gradient differences for vanilla pooling methods 
As softmax is differentiable, unlike maximum or stochastic pooling methods, during backpropagation, a minimum non-zero weight will be assigned to every positive activation within a kernel region. This enables the calculation of a gradient for every non-zero activation in that region, as shown in Figure~\ref{fig:softpool}.

% Dealing with finite ranges
In our implementation of SoftPool, we use finite ranges of possible values given a precision level (i.e., half, single or double) as detailed in \textsection \textcolor{red}{7} of the Supplementary Material. We retain the differentiable nature of softmax by assigning a lower arithmetic limit given the number of bits used by each type preventing arithmetic underflow.

\subsection{Feature preservation}
\label{sec:methodology::feature_preservation}

% Importance of feature preservation
An integral goal of sub-sampling is the preservation of  representative features in the input, while simultaneously minimizing the overall resolution. Creating unrepresentative downsampled versions of the original inputs can be harmful to the overall model's performance as the representation of the input is detrimental for the task.

\begin{figure*}[ht]
\centering
\begin{overpic}[width=\textwidth]{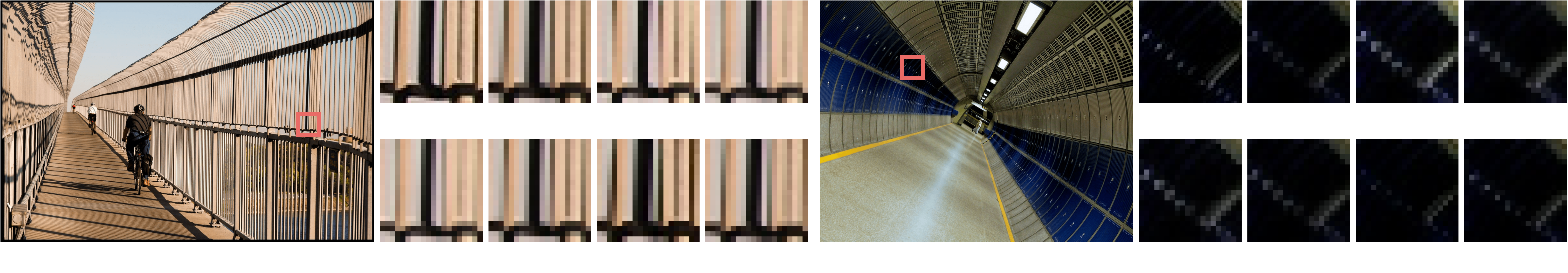}
%grid,tics=5
\put (24.7,9.7) {\tiny{Original (HR)}}
\put (33.1,9.7) {\tiny{Average}}
\put (39.4,9.7) {\tiny{Maximum}}
\put (47.0,9.7) {\tiny{S3}\cite{zhai2017s3pool}}

\put (24.5,0.8) {\tiny{Stochastic} \cite{zeiler2013stochastic}}
\put (32.5,0.8) {\tiny{$L_{p}$} \cite{gulcehre2014learned}}
\put (39.5,0.8) {\tiny{Gate} \cite{lee2016generalizing}}
\put (45.15,0.8) {\tiny{SoftPool \textbf{(ours)}}}

\put (73.1,9.7) {\tiny{Original (HR)}}
\put (81.5,9.7) {\tiny{Average}}
\put (87.8,9.7) {\tiny{Maximum}}
\put (95.4,9.7) {\tiny{S3}\cite{zhai2017s3pool}}

\put (72.9,0.8) {\tiny{Stochastic} \cite{zeiler2013stochastic}}
\put (80.9,0.8) {\tiny{$L_{p}$} \cite{gulcehre2014learned}}
\put (87.9,0.8) {\tiny{Gate} \cite{lee2016generalizing}}
\put (93.55,0.8) {\tiny{SoftPool \textbf{(ours)}}}

\end{overpic}
\vspace{-4mm}
\caption{ \textbf{Visual comparisons over pooling methods}. Both images are from Urban100 \cite{huang2015single}. Left image (\textit{img024}) shows high contrasting borders/edges. Right image (\textit{img078}) presents the inverse of high local values within an overall low-valued region.}
\label{fig:sim_tests}
\vspace{-.5em}
\end{figure*}
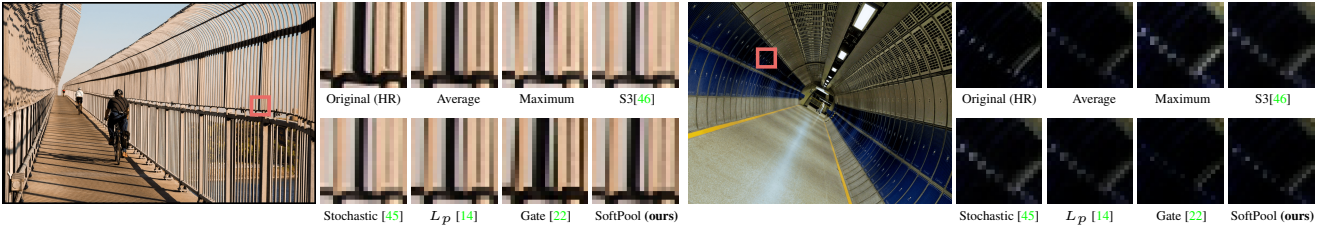

% Detail with SoftPool
Currently widely used pooling techniques can be ineffective in certain cases, as shown in Figures~\ref{fig:softpool_examples} \& \ref{fig:sim_tests}. Average pooling decreases the effect of all activations in the region equally, while max pooling selects only the single highest activation in the region. SoftPool falls between the two, as all activations in the region contribute to the final output, with higher activations are more dominant than lower ones. This balances the effects of both average and max pooling, while leveraging the beneficial properties of both. 

\subsection{Spatio-temporal kernels}
\label{sec:methodology::spacetime}
% Extension to space-time data
CNNs have also been extended to 3D inputs to include additional dimensions such as depth and time. To accommodate these inputs, we extend SoftPool to include an additional dimension. For an input activation map $\mathbf{a}$ of $C \times H \times W \times T$, with $T$ the temporal extent, we transform the 2D spatial kernel region $\textbf{R}$ to a 3D spatio-temporal region with an additional third temporal dimension.

% Challenges with space-time data
The produced output holds condensed spatio-temporal information. Issues that arise with the introduction of the temporal dimension are discussed and illustrated in \textsection \textcolor{red}{3} of the Supplementary Material. With the added dimension, desired pooling properties such as limited loss of information, a differentiable function, and low computational and memory overhead are even more important. %In the next section, we demonstrate that SoftPool performs favorably with 3D inputs.

\begin{table*}[t]
\centering
\resizebox{\textwidth}{!}{%
\begin{tabular}{cl|cc|cc|cc|cc|cc|cc|cc|cc|cc}
\hline
&
\multirow{3}{*}{Pooling method} &
\multicolumn{6}{c|}{DIV2K \cite{agustsson2017ntire}} &
\multicolumn{6}{c|}{Urban100 \cite{huang2015single}} & 
\multicolumn{6}{c}{Manga109 \cite{matsui2017sketch}} \\[.3em]
& &
\multicolumn{2}{c|}{$k=2$} &
\multicolumn{2}{c|}{$k=3$} &
\multicolumn{2}{c|}{$k=5$} &
\multicolumn{2}{c|}{$k=2$} &
\multicolumn{2}{c|}{$k=3$} &
\multicolumn{2}{c|}{$k=5$} &
\multicolumn{2}{c|}{$k=2$} &
\multicolumn{2}{c|}{$k=3$} &
\multicolumn{2}{c}{$k=5$} \\[.2em]\cline{3-20}
& 

& SSIM & PSNR 
& SSIM & PSNR 
& SSIM & PSNR

& SSIM & PSNR 
& SSIM & PSNR 
& SSIM & PSNR 

& SSIM & PSNR 
& SSIM & PSNR 
& SSIM & PSNR \\[.2em]
\hline
\parbox[t]{2mm}{\multirow{3}{*}{\rotatebox[origin=c]{90}{}}} 

& Average  
& 0.714 & 51.247
& 0.578 & 44.704
& 0.417 & 29.223
& 0.691 & 50.380
& 0.563 & 41.745
& 0.372 & 28.270
& 0.695 & 54.326
& 0.582 & 43.657
& 0.396 & 29.862 \\[.15em]

& Maximum 
& 0.685 & 49.826
& 0.370 & 41.944
& 0.358 & 22.041
& 0.662 & 48.266
& 0.528 & 40.709
& 0.330 & 20.654 
& 0.671 & 50.085
& 0.544 & 41.128
& 0.324 & 22.307 \\[.15em]

& Pow-average
& 0.419 & 35.587
& 0.286 & 26.329
& 0.178 & 16.567
& 0.312 & 31.911
& 0.219 & 24.698
& 0.124 & 15.659 
& 0.381 & 29.248
& 0.276 & 18.874
& 0.160 & 9.266 \\[.15em]

& Sum
& 0.408 & 35.153
& 0.268 & 26.172
& 0.193 & 17.315
& 0.301 & 31.657
& 0.208 & 24.735
& 0.123 & 15.243 
& 0.374 & 30.169
& 0.271 & 20.150
& 0.168 & 13.081 \\[.15em]\cline{1-20}

\parbox[t]{2mm}{\multirow{4}{*}{\rotatebox[origin=c]{90}{Trainable}}} 
& $L_{p}$ \cite{gulcehre2014learned}
& 0.686 & 49.912
& 0.542 & 43.083
& 0.347 & 25.139
& 0.676 & 48.508
& 0.534 & 39.986
& 0.326 & 26.365 
& 0.675 & 51.721
& 0.561 & 41.824
& 0.367 & 27.469 \\[.15em]

& Gate \cite{lee2016generalizing} 
& 0.689 & 50.104
& 0.560 & 43.437
& 0.353 & 25.672 
& 0.675 & 49.769
& 0.537 & 40.422
& 0.328 & 26.731 
& 0.679 & 51.980
& 0.569 & 42.127
& 0.374 & 27.754 \\[.15em]

& Lite-S3DPP \cite{saeedan2018detail}  
& 0.702 & 50.598
& 0.562 & 44.076
& 0.396 & 27.421 
& 0.684 & 49.947
& 0.551 & 40.813
& 0.365 & 27.136 
& 0.691 & 52.646
& 0.573 & 42.794
& 0.386 & 28.598 \\[.15em]

& LIP \cite{gao2019lip}
& 0.711 & 50.831
& 0.559 & 44.432
& 0.401 & 28.285 
& 0.689 & 50.266
& 0.558 & 41.159
& 0.370 & 27.849 
& 0.689 & 53.537
& 0.579 & 43.018
& 0.391 & 29.331 \\[.15em]\cline{1-20}

\parbox[t]{2mm}{\multirow{2}{*}{\rotatebox[origin=c]{90}{Stoch.}}} 
& Stochastic \cite{zeiler2013stochastic} 
& 0.631 & 45.362
& 0.479 & 39.895
& 0.295 & 21.314
& 0.616 & 44.342
& 0.463 & 37.223
& 0.286 & 19.358 
& 0.583 & 46.274
& 0.427 & 39.259
& 0.255 & 22.953 \\[.3em]

& S3 \cite{zhai2017s3pool} 
& 0.609 & 44.760
& 0.454 & 39.326
& 0.280 & 20.773 
& 0.608 & 44.239
& 0.459 & 36.965
& 0.272 & 19.645
& 0.576 & 46.613 
& 0.426 & 39.866
& 0.232 & 23.242 \\[.3em]

\cline{1-20}

\parbox[t]{2mm}{\rotatebox[origin=c]{90}{~}} 
& SoftPool (Ours)  
& \textbf{0.729} & \textbf{51.436}
& \textbf{0.594} & \textbf{44.747}
& \textbf{0.421} & \textbf{29.583}
& \textbf{0.694} & \textbf{50.687}
& \textbf{0.578} & \textbf{41.851}
& \textbf{0.394} & \textbf{28.326} 
& \textbf{0.704} & \textbf{54.563}
& \textbf{0.586} & \textbf{43.782}
& \textbf{0.403} & \textbf{30.114} \\[.15em]
\end{tabular}
}
\caption{\textbf{Quantitative results on benchmark high-res datasets}. Best results from each similarity measure are denoted in \textbf{bold}.} 
\label{tab:times_ssim_psnr}
\vspace{-1em}
\end{table*}

\section{Experimental Results} \label{sec:experiments}

We first evaluate the information loss for various pooling operators. We compare the downsampled outputs to the original inputs using standard similarity measures (Section~\ref{sec:experiments::similarity}). We also investigate each pooling operator's computation and memory overhead (Section~\ref{sec:experiments::latency}).

We then focus on the classification performance gain when using SoftPool in a range of popular CNN architectures and in comparison to other pooling methods (Section~\ref{sec:experiments::imagenet}). We also perform an ablation study where we replace max-pooling operations by SoftPool operations in an InceptionV3 architecture (Section~\ref{sec:experiments::multi-layer}).

Finally, we demonstrate the merits of SoftPool for spatio-temporal data by focusing on action recognition in video (Section~\ref{sec:experiments::video}). Additionally, we investigate how transfer learning performance is affected when SoftPool is used.

\subsection{Experimental settings}
\label{sec:experiments::settings}

% datasets for image/video recognition
\textbf{Datasets}. For our experiments with images, we employ five different datasets for the tasks of image quality quantitative evaluation and classification. For image quality and similarity assessment we use high-resolution DIV2K \cite{agustsson2017ntire}, Urban100 \cite{huang2015single}, Manga109 \cite{matsui2017sketch}, and Flicker2K \cite{agustsson2017ntire}. ImageNet1K \cite{ILSVRC15} is used for the classification task. For video-based action recognition, we use the large-scale HACS \cite{zhao2019hacs} and Kinetics-700 \cite{carreira2019short} datasets. Our transfer learning experiments are performed on the UCF-101 \cite{soomro2012ucf101} dataset.

% image-based tests settings
\textbf{Training implementation details}. For the image classification task, we perform random region selection of $294 \times 294$ height and width, which was then resized to $224 \times 224$. We use an initial learning rate of 0.1 with an SGD optimizer and a step-wise learning rate reduction every 40 epochs for a total of 100 epochs. The epochs number was chosen as no further improvements were observed for any of the models. We also set the batch size to 256 across all models.

% Video-based tests settings
For our experiments on videos, we use a multigrid training scheme \cite{wu2020multigrid}, with frame sizes between 4--16 and frame crops of 90--256 depending on the cycle. On average, the video inputs are of size $8 \times 160 \times 160$. With the multigrid scheme, the batch sizes are between 64 and 2048 with each size counter-equal to the input size in every step to match the memory use. We use the same learning rate, optimizer, learning rate schedule and maximum number of epochs as in the image-based experiments.

\begin{table*}[t]
\centering
\resizebox{\linewidth}{!}{%
\begin{tabular}{l|c|c|cc|cc|cc}
\hline
\multirow{2}{*}{Model} & Params & \multirow{2}{*}{GFLOP} & \multicolumn{2}{c|}{Original} & \multicolumn{2}{c|}{SoftPool (pre-train)} & \multicolumn{2}{c}{SoftPool (from scratch)} \\
& (M) & & top-1 & top-5 & top-1 & top-5 & top-1 & top-5 \\
\hline
ResNet18 & 11.7 & 1.83 & 
69.76 & 
89.08 & 
70.56 (\textcolor{applegreen}{+0.80}) & 
89.89  (\textcolor{applegreen}{+0.81}) & 
\textbf{71.27} (\textcolor{applegreen}{+1.51}) & 
\textbf{90.16} (\textcolor{applegreen}{+1.08})\\[0.1em]

ResNet34 & 21.8 & 3.68 & 
73.30 & 
91.42 & 
74.03 (\textcolor{applegreen}{+0.73}) & 
91.85 (\textcolor{applegreen}{+0.43}) &
\textbf{74.67} (\textcolor{applegreen}{+1.37}) & 
\textbf{92.30} (\textcolor{applegreen}{+0.88})\\[0.1em]

ResNet50 & 25.6 & 4.14 & 
76.15 & 
92.87 & 
76.60 (\textcolor{applegreen}{+0.45}) & 
93.15 (\textcolor{applegreen}{+0.28}) &
\textbf{77.35} (\textcolor{applegreen}{+1.17}) & 
\textbf{93.63} (\textcolor{applegreen}{+0.76})\\[0.1em]

ResNet101 & 44.5 & 7.87 & 
77.37 & 
93.56 & 
77.74 (\textcolor{applegreen}{+0.37}) & 
93.99 (\textcolor{applegreen}{+0.43}) &
\textbf{78.32} (\textcolor{applegreen}{+0.95}) & 
\textbf{94.21} (\textcolor{applegreen}{+0.65})\\[0.1em]

ResNet152 & 60.2 & 11.61 & 
78.31 & 
94.06 & 
78.73 (\textcolor{applegreen}{+0.42}) & 
94.47 (\textcolor{applegreen}{+0.41}) & 
\textbf{79.24} (\textcolor{applegreen}{+0.92}) & 
\textbf{94.72} (\textcolor{applegreen}{+0.66})\\[0.1em]

\hline
DenseNet121 & 8.0 & 2.90 & 
74.65 & 
92.17 & 
75.27 (\textcolor{applegreen}{+0.57}) & 
92.60 (\textcolor{applegreen}{+0.43}) & 
\textbf{75.88} (\textcolor{applegreen}{+1.23}) & 
\textbf{92.92} (\textcolor{applegreen}{+0.75})\\[0.1em]

DenseNet161 & 28.7 & 7.85 & 
77.65 & 
93.80 & 
78.12 (\textcolor{applegreen}{+0.47}) & 
94.15 (\textcolor{applegreen}{+0.35}) & 
\textbf{78.72} (\textcolor{applegreen}{+0.93}) & 
\textbf{94.41} (\textcolor{applegreen}{+0.61})\\[0.1em]

DenseNet169 & 14.1 & 3.44 & 
76.00 & 
93.00 & 
76.49 (\textcolor{applegreen}{+0.49}) & 
93.38 (\textcolor{applegreen}{+0.38}) &
\textbf{76.95} (\textcolor{applegreen}{+0.95}) & 
\textbf{93.76} (\textcolor{applegreen}{+0.76})\\[0.1em]
\hline

ResNeXt50 32x4d & 25.0 & 4.29 & 
77.62 & 
93.70 & 
78.23 (\textcolor{applegreen}{+0.61}) &
93.97 (\textcolor{applegreen}{+0.27}) & 
\textbf{78.48} (\textcolor{applegreen}{+0.86}) & 
\textbf{93.37} (\textcolor{applegreen}{+0.67})\\[0.1em]

ResNeXt101 32x8d & 88.8 & 7.89 & 
79.31 & 
94.28 & 
78.89 (\textcolor{applegreen}{+0.58}) & 
94.73 (\textcolor{applegreen}{+0.45}) & 
\textbf{80.12} (\textcolor{applegreen}{+0.81}) & 
\textbf{94.88} (\textcolor{applegreen}{+0.60})\\[0.1em]

\hline
Wide-ResNet50 & 68.9 & 11.46 & 
78.51 & 
94.09 & 
79.14 (\textcolor{applegreen}{+0.63}) & 
94.51 (\textcolor{applegreen}{+0.42}) & 
\textbf{79.52} (\textcolor{applegreen}{+1.01}) & 
\textbf{94.85} (\textcolor{applegreen}{+0.76})\\[0.1em]

\end{tabular}
}
\caption{\textbf{Trained from scratch and pre-trained pairwise comparisons of top-1 and top-5 accuracies} on ImageNet1K between the original networks and the same networks with pooling replaced by SoftPool.}
\label{tab:ImageNet_pre_retrained}
\vspace{-1em}
\end{table*}

\subsection{Downsampling similarity}
\label{sec:experiments::similarity}

% Similarity tests
We first assess the information loss of various pooling operations. We compare the original inputs with the downsampled outputs in terms of similarity. We use three of the most widely used kernel sizes (i.e., $k \! \times \! k$, with $k = \{2,3,5\})$. Our experiments are based on two standardized image similarity evaluation metrics \cite{wang2004image}:

% Structural Similarity Index
\textbf{Structural Similarity Index Measure (SSIM)} is used between the original and downsampled images. SSIM is based on the computation of a luminance, contrast and structural term. Larger index values correspond to larger structural similarities between the images compared.

% Pixel-wise similarity
\textbf{Peak Signal-to-Noise Ratio (PSNR)} measures the compression quality of the resulting image based on the Mean Squared Error (MSE) inverse between the weighted averages of their channels. PSNR depends on the MSE with higher values relate to lower errors between the two images.

\begin{figure}[t]
\centering
\includegraphics[width=\linewidth]{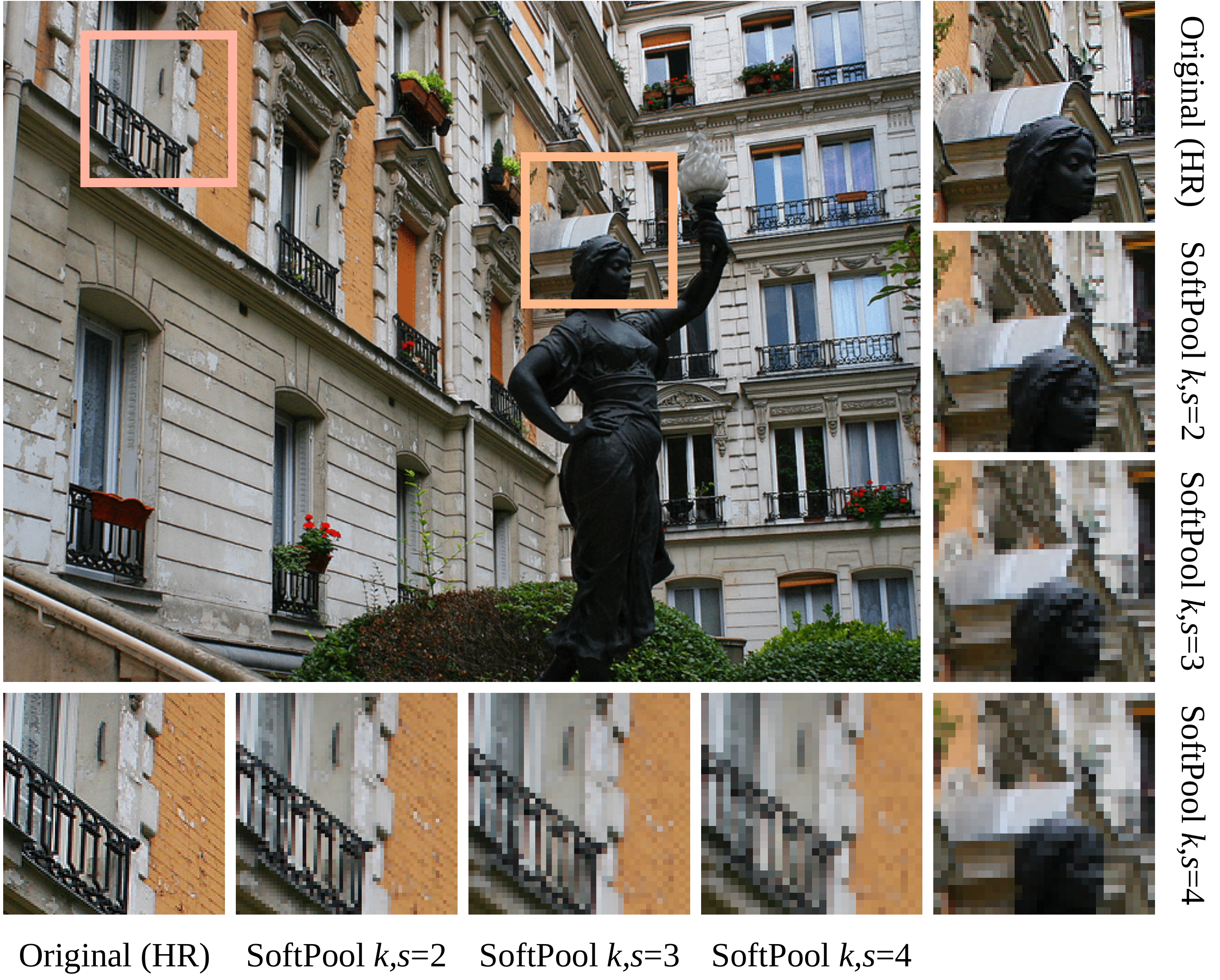}
\caption{\textbf{Results after SoftPool pooling with different kernel and stride sizes}. Image (\textit{img003}) is from Urban100 \cite{huang2015single}.}
\label{fig:softpool_op}
\vspace{-1.2em}
\end{figure}

% visual example
Visual examples of different compression methods are shown in Figure~\ref{fig:sim_tests}. The proposed SoftPool method can represent regions with borders between low and high frequencies better than other methods, shown by the border of the black bar in the left image. The inverse also hold true for a high-frequency location within an overall low-frequency region as shown in the left image. In such cases, max and stochastic-based methods \cite{gulcehre2014learned,zeiler2013stochastic,zhai2017s3pool} over-amplify pixel locations in the subsampled volumes while these pixels are completely lost in the downsampled volume through average and gate \cite{lee2016generalizing} cases. In contrast, SoftPool shows the ability to preserve such patterns as also shown in Figure~\ref{fig:softpool_op}.

% SSI and PSNR of different pooling methods
In Tables~\ref{tab:times_ssim_psnr} and \ref{tab:compute_flicker}, we show the average SSIM and PSNR values obtained on DIV2K \cite{agustsson2017ntire}, Urban100 \cite{huang2015single}, Manga109 \cite{matsui2017sketch}, and Flicker2K \cite{agustsson2017ntire} high-resolution datasets over different kernel sizes. For both measures, SoftPool outperforms all other methods by a reasonable margin. Notably, it significantly outperforms non-trainable and stochastic methods. The randomized strategy of stochastic methods does not effectively allow their use as a standalone method as they lack non-linear operations. Trainable approaches are bounded by both the image types they have been trained on as well as on the discovered channel correlations during pooling. 

\begin{table}[t]
\centering
\resizebox{\linewidth}{!}{%
\begin{tabular}{l|c|c|cc|cc|cc}
\hline
\multirow{3}{*}{Pooling} &
\multirow{2}{*}{CPU (ms)} &
\multirow{2}{*}{CUDA (ms)} &
\multicolumn{6}{c}{Flicker2K\cite{agustsson2017ntire}} \\[.3em]
 & & &   
\multicolumn{2}{c|}{$k=2$} &
\multicolumn{2}{c|}{$k=3$} &
\multicolumn{2}{c}{$k=5$} \\[.2em]\cline{4-9}
& ($\downarrow$ F / $\uparrow$ B) & ($\downarrow$ F / $\uparrow$ B)
& SSIM & PSNR 
& SSIM & PSNR 
& SSIM & PSNR\\[.2em]
\hline

Avg & 9 / 49 & 14 / 76 
& 0.709 & 51.786
& 0.572 & 44.246
& 0.408 & 28.957\\[.15em]

Max & 91 / 152 & 195 / 267 
& 0.674 & 47.613
& 0.385 & 40.735
& 0.329 & 21.368 \\[.15em]

Pow-avg & 74 / 329 & 120 / 433 
& 0.392 & 34.319
& 0.271 & 26.820
& 0.163 & 15.453\\[.15em]

Sum & 26 / 163 & 79 / 323 
& 0.386 & 34.173
& 0.265 & 26.259
& 0.161 & 15.218\\[.15em]\cline{1-9}

$L_{p}$ \cite{gulcehre2014learned} & 116 / 338 & 214 / 422
& 0.683 & 48.617
& 0.437 & 42.079
& 0.341 & 24.432\\[.15em]

Gate \cite{lee2016generalizing} & 245 / 339 & 327 / 540 
& 0.687 & 49.314
& 0.449 & 42.722
& 0.358 & 25.687 \\[.15em]

DPP \cite{saeedan2018detail}  & 427 / 860 & 634 / 1228
& 0.691 & 50.586
& 0.534 & 43.608
& 0.385 & 27.430 \\[.15em]

LIP \cite{gao2019lip} & 134 / 257 & 258 / 362  
& 0.696 & 50.947
& 0.548 & 43.882
& 0.390 & 28.134 \\[.15em]\cline{1-9}

Stoch. \cite{zeiler2013stochastic} & 162 / 341 & 219 / 485 
& 0.625 & 46.714
& 0.474 & 38.365
& 0.264 & 21.428 \\[.3em]

S3 \cite{zhai2017s3pool} & 233 / 410 & 345 / 486  
& 0.611 & 46.547
& 0.476 & 37.706
& 0.252 & 21.363 \\[.3em]

\cline{1-9}

SoftPool & 31 / 156 & 56 / 234 
& \textbf{0.721} & \textbf{52.356}
& \textbf{0.587} & \textbf{44.893}
& \textbf{0.416} & \textbf{29.341} \\[.15em]

\end{tabular}
}
\caption{\textbf{Latency and pixel similarity}. Latency times are averaged over ImageNet1K \cite{ILSVRC15}.\vspace{-0.2em}}
\label{tab:compute_flicker}
\vspace{-1.1em}
\end{table}

\subsection{Latency and memory use} 
\label{sec:experiments::latency}
Memory and latency costs of pooling operations are largely overlooked as a single operation has negligible latency times and memory consumption. However, because of the parallelization of deep learning models, operations may be performed thousands of times per step. Eventually, a slow or memory-intensive pooling operation can have a detrimental effect on the performance.

% Inference time
To test the computation and memory overhead, we report running-time memory use and inference on both CPU and GPU (CUDA) in Table~\ref{tab:compute_flicker}. We detail our testing environment and implementation in \textsection \textcolor{red}{7} of the Supplementary Material. 

% latency/memory results 
From Table~\ref{tab:compute_flicker} we observe that our implementation of SoftPool achieves low inference times for both CPU- and CUDA-based operations, while remaining memory-efficient. This is because the method allows parallelization and is simple to compute within a region. SoftPool is second only to average pooling in terms of latency and memory use as operations can be performed in-place at the tensor.

\subsection{Classification performance on ImageNet1K}
\label{sec:experiments::imagenet}

% Layer-wise pooling replacements
We investigate whether classification accuracy improves as a result of SoftPool's superior ability to retain information. We replace the original pooling layers in ResNet \cite{he2016deep}, DenseNet \cite{huang2017densely}, ResNeXt \cite{xie2017aggregated} and wide-ResNet \cite{zagoruyko2016wide} networks. These models were chosen because of their wide use. We consider two distinct settings. In the \textit{from scratch} setting, we replace the pooling operators of the original models by SoftPool and train with weights randomly initialized. In the \textit{pre-trained} setting, we replace pooling layers of the original trained networks and evaluate the effects of the layer change on the ImageNet1K validation set without further training. Results of both settings appear in Table~\ref{tab:ImageNet_pre_retrained}.

% results
Networks trained from scratch with pooling layers replaced by SoftPool yield consistent accuracy improvements over the original networks. The same trend is also visible for the pre-trained networks for which the models have been trained with their original pooling methods. We now discuss the results per CNN architecture family.

% ResNets
\textbf{ResNet} \cite{he2016deep}. By training from scratch with SoftPool, an average top-1 accuracy improvement of 1.17\% is observed with a maximum of 1.51\% on ResNet18. When replacing pooling layers on pre-trained networks, we obtain an average of +0.59\% accuracy. All ResNet-based models only include a single pooling operation after the first convolution. Thus, results are based on a single layer replacement which emphasizes the merits of using SoftPool.

\begin{table}[t]
\centering
\resizebox{\linewidth}{!}{%
\begin{tabular}{l|c|cc}
\hline
Model & pooling replacement & statistic ($\chi ^{2}$) & p-value ($\rho$)\\[.2em]
\hline
ResNet18 & \multirow{5}{*}{Max $\rightarrow$ SoftPool} & $90.08$ & $2.28e^{-21}$ \\[.15em]
ResNet34 & & $15.04$ & $1.05e^{-4}$ \\[.15em]
ResNet50 & & $61.80$ & $3.80e^{-15}$ \\[.15em]
ResNet101 & & $23.13$ & $1.51e^{-6}$ \\[.15em]
ResNet152 & & $50.29$ & $1.32e^{-12}$ \\[.15em]
\hline
DenseNet121 & \multirow{3}{*}{Avg+Max $\rightarrow$ SoftPool} & $411.02$ & $2.19^{-91}$\\[.15em]
DenseNet161 &  & $46.52$ & $9.06e^{-12}$ \\[.15em]
DenseNet169 &  & $27.32$ & $1.72e^{-7}$ \\[.15em]
\hline
ResNeXt50 32x4d & \multirow{2}{*}{Max $\rightarrow$ SoftPool} & $45.13$ & $1.85e^{-11}$ \\[.15em]
ResNeXt101 32x8d &  & $695.87$ & $2.36e^{-53}$ \\[.15em]
\hline
wide-ResNet50 & Max $\rightarrow$ SoftPool & $39.41$ & $3.43e^{-10}$ \\[.15em]
\hline
InceptionV1 & \multirow{2}{*}{Max $\rightarrow$ SoftPool} & $106.41$ & $5.98e^{-25}$ \\[.15em]
InceptionV3 &  & $80.19$ & $3.39e^{-19}$ \\[.15em]

\end{tabular}
}
\caption{\textbf{McNemar's} \cite{edwards1948note,mcnemar1947note} \textbf{statistical significance for pooling substitution results correlation}. The top-1 accuracies on ImageNet1K val set from Tables~\ref{tab:ImageNet_pre_retrained}, \ref{tab:pooling_imagenet_tests} and \ref{tab:inception_tests_progressive} are used. A detailed view of the statistics is presented in \textsection \textcolor{red}{6} in the Supplementary Material.}
\label{tab:mcnemar_stat_sign}
\vspace{-0.5em}
\end{table}

% DenseNets
\textbf{DenseNet} \cite{huang2017densely}. Based on the DenseNet overall architecture that incorporates five pooling operations, we replace the max pooling operation that follows after the first layer and average pooling layers between Dense blocks with our proposed method. Top-1 accuracy gains are in the 0.93--1.23\% range when training from scratch and 0.47--0.57\% with substitution on pre-trained networks. The top-5 accuracy follows similar incremental trends. The largest increment in both settings is observed for DenseNet121.

% ResNeXts
\textbf{ResNeXt} \cite{xie2017aggregated}. Average of 0.83\% top-1 accuracy improvement when trained from scratch. The best model, ResNeXt101 32x8d, achieves 80.12\% top-1 accuracy (+0.81\%) and 94.88\% top-5 accuracy (+0.60\%) with SoftPool. On the pre-trained settings, we note average improvements of +0.59\% on top-1 and +0.36\% top-5 accuracies. We again note these accuracies come by a replacement of their only pooling operation after the first convolution layer and without additional training.

% Wide-ResNets
\textbf{Wide-Resnet-50} \cite{zagoruyko2016wide}. We observe top-1 and top-5 accuracy increases of 1.01\% and 0.76\% when trained from scratch with SoftPool. The initialized network also achieves improvements with +0.63\% top-1 and +0.42\% top-5.

\begin{table}[t]
\centering
\resizebox{\linewidth}{!}{%
\begin{tabular}{lccccccc}
\hline
\multirow{2}{*}{Pooling} & \multicolumn{7}{c}{Networks} \\[0.2em]\cline{2-8}
\rule{0pt}{50pt} 
& \rotatebox{35}{\parbox{2mm}{ResNet18\;\cite{he2016deep}}} 
& \rotatebox{35}{\parbox{2mm}{ResNet34\;\cite{he2016deep}}} 
& \rotatebox{35}{\parbox{2mm}{ResNet50\;\cite{he2016deep}}} 
& \rotatebox{35}{\parbox{2mm}{ResNeXt50\;\cite{xie2017aggregated}}}
& \rotatebox{35}{\parbox{2mm}{DenseNet121\;\cite{huang2017densely}}} 
& \rotatebox{35}{\parbox{2mm}{InceptionV1\;\cite{szegedy2015going}}}
& \\[0.2em]\cline{1-7}
\multirow{2}{*}{Original} & (Max) & (Max) & (Max) & 
(Max) & (Avg+Max) & (Max) & \\[0.1em]
& 69.76 & 73.30 & 76.15 & 
77.62 & 74.65 & 69.78 &  \\[0.2em]\cline{1-7}

Stochastic \cite{zeiler2013stochastic} & 70.13 & 73.34 & 76.11 &
77.71 & 74.84 & 70.14 & \\[0.1em]

S3 \cite{zhai2017s3pool} & 70.15 & 73.56 & 76.24 &
77.82 & 74.85 & 70.17 & \\[0.2em]\cline{1-7}

$L_{p}$ \cite{lee2016generalizing} & 70.45 & 73.74 & 76.56 &
77.86 & 74.93 & 70.32 & \\[0.1em]

Gate \cite{gulcehre2014learned} & 70.74 & 73.68 & 76.75 &
77.98 & 74.88 & 70.52 & \\[0.1em]

DPP \cite{saeedan2018detail} & 70.86 & 74.25 & 77.09 &
78.20 & 75.37 & 70.95 & \\[0.1em]

LIP \cite{gao2019lip} & 70.83 & 73.95 & 77.13 &
78.14 & 75.31 & 70.77 & \\[0.2em]\cline{1-7}

SoftPool \textbf{(ours)} & \textbf{71.27} & \textbf{74.67} & \textbf{77.35} & \textbf{78.48} & \textbf{75.88} & \textbf{71.43} & \:\:\:\:\:\:\: \\

\end{tabular}
}
\caption{\textbf{Pooling layer substitution} top-1 accuracy for a variety of pooling methods. Experiments were performed on Imagenet1K.}
\label{tab:pooling_imagenet_tests}
\vspace{-1.2em}
\end{table}

\begin{table}[t]
\centering
\resizebox{\linewidth}{!}{%
\begin{tabular}{lcccccccc}
\hline
\multirow{2}{*}{Layer} & \multicolumn{8}{c}{Pooling layer substitution with SoftPool} \\\cline{2-9}
& N & I & II & III & IV & V & VI & VII\\
\hline
$pool_{1}$ & & \checkmark & \checkmark & \checkmark & \checkmark & \checkmark & \checkmark & \checkmark \\ 
$pool_{2}$ & & & \checkmark & \checkmark & \checkmark & \checkmark & \checkmark & \checkmark \\
$mixed \; 5_{b-d}$ & & & & \checkmark & \checkmark & \checkmark & \checkmark & \checkmark \\
$mixed \; 6_{a}$ & & & & & \checkmark & \checkmark & \checkmark & \checkmark \\
$mixed \; 6_{b-e}$ & & & & & & \checkmark & \checkmark & \checkmark \\
$mixed \; 7_{a}$ & & & & & & & \checkmark & \checkmark \\
$mixed \; 7_{b-d}$ & & & & & & & & \checkmark \\
\hline
Top-1 (\%) & 77.45 & 77.93 & 78.14 & 78.37 & 78.42 & 78.65 & 78.83 & \textbf{79.04} \\
Top-5 (\%) & 93.56 & 93.61 & 93.68 & 93.74 & 93.78 & 93.84 & 93.90 & \textbf{93.98} \\
\end{tabular}
}
\caption{\textbf{Progressive layer substitution for InceptionV3}. Experiments are performed on ImageNet1K. Column numbers refer to the number of replaced pooling layers, marked with \checkmark.}
\label{tab:inception_tests_progressive}
\vspace{-1.2em}
\end{table}

\begin{table*}[t]
\begin{threeparttable}[t]
\centering
\resizebox{\textwidth}{!}{%
\begin{tabular}{lccccccc}
\hline
\multirow{2}{*}{Model} & 
\multirow{2}{*}{GFLOPs} &
\multicolumn{2}{c}{HACS} &
\multicolumn{2}{c}{Kinetics-700} &
\multicolumn{2}{c}{UCF-101} \\
&
&
top-1(\%) & top-5(\%) & 
top-1(\%) & top-5(\%) &
top-1(\%) & top-5(\%) \\
\hline
r3d-50 \cite{kataoka2020would}$^{**}$ &
53.16 &
78.36 & 93.76 &
49.08 & 72.54 & 
93.13 & 96.29 \\
r3d-101 \cite{kataoka2020would}$^{**}$ &
78.52 &
80.49 & 95.18 &
52.58 & 74.63 & 
95.76 & 98.42 \\
r(2+1)d-50 \cite{tran2018closer}$^{**}$ &
50.04 &
81.34 & 94.51 &
49.93 & 73.40 & 
93.92 & 97.84 \\
I3D \cite{carreira2017quo}$^{\ddagger *}$ &
55.27 &
79.95 & 94.48 &
53.01 & 69.19 & 
92.45 & 97.62 \\
ir-CSN-101 \cite{tran2019video}$^{\ddagger \dagger}$ &
17.26 &
N/A & N/A &
54.66 & 73.78 &
95.13 & 97.85 \\
MF-Net \cite{chen2018multifiber}$^{\dagger *}$ &
22.50 &
78.31 & 94.62 &
54.25 & 73.38 &
93.86 & 98.37 \\
SlowFast r3d-50 \cite{feichtenhofer2019slowfast}$^{\ddagger \dagger}$ &
36.71 &
N/A & N/A &
56.17 & 75.57 &
94.62 & 98.75 \\
SRTG r3d-50 \cite{stergiou2021learn}$^{\dagger \dagger}$ &
53.22 &
80.36 & 95.55 &
53.52 & 74.17 & 
96.85 & 98.26 \\
SRTG r(2+1)d-50 \cite{stergiou2021learn}$^{\dagger \dagger}$ &
50.10 &
83.77 & 96.56 &
54.17 & 74.62 &
95.99 & 98.20 \\
SRTG r3d-101 \cite{stergiou2021learn}$^{\dagger \dagger}$ &
78.66 &
81.66 & 96.37 &
56.46 & 76.82 &
97.32 & 99.56 \\
\hline
r3d-50 with SoftPool (Ours) &
53.16 &
79.82 & 94.64 &
50.36 & 73.72 &
93.90 & 97.02 \\
SRTG r(2+1)d-50 with SoftPool (Ours) &
50.10 &
\textbf{84.78} & \textbf{97.72} &
55.27 & 75.44 &
96.46 & 98.73 \\
SRTG r3d-101 with SoftPool (Ours) &
78.66 &
83.28 & 97.04 &
\textbf{57.76} & \textbf{77.84} &
\textbf{98.06} & \textbf{99.82} \\
\hline
\end{tabular}%
}
 \begin{tablenotes}
    \item[$**$] re-implemented models trained from scratch.  $\dagger \dagger$ models and weights from official repositories. $\ddagger *$ unofficial models trained from scratch.
    \item[$\ddagger \dagger$] models from unofficial repositories with official weights.  $\dagger *$ official models trained from scratch.
   \end{tablenotes}
\end{threeparttable}%
\vspace{0.8mm}
\caption{\textbf{Action recognition top-1 and top-5 accuracy for HACS, Kinetics-700 and UCF-101}. Models are trained on HACS and fine-tuned for Kinetics-700 and UCF-101, except for ir-CSN-101 and SlowFast r3d-50 (see text). N/A means no trained model was provided.
\label{table:video_accuracies}
}
\vspace{-1.3em}
\end{table*}

% conclusion
These combined experiments demonstrate that by replacing a single pooling layer (ResNet, ResNeXt and Wide-ResNet) or only five pooling layers (DenseNet) with SoftPool operations leads to a modest but important increase in accuracy. To understand whether these improvements are statistically significant, we performed a McNemar's test \cite{edwards1948note,mcnemar1947note} to calculate the probabilities ($\rho$) of marginal homogeneity between the original and the SoftPool-replaced networks. The results are summarized in Table~\ref{tab:mcnemar_stat_sign} for the models obtained in the \textit{from scratch} setting. As shown for all networks, $\rho \! \ll \! 0.01$ which corresponds to confidence $\gg \! 99\%$ that the improved results are indeed due to the different pooling operations.

% memory and GFLOPs
\textbf{Memory and computation requirements}. We also summarize the number of parameters and GLOPs in Table~\ref{tab:ImageNet_pre_retrained}. SoftPool does not include trainable variables and thus does not affect the number of parameters, in contrast to recent pooling methods \cite{gao2019lip,gulcehre2014learned,kobayashi2019global,saeedan2018detail}. We also depart from these methods as the number of GFLOPs remains the same as the maximum and average pooling that we replace.

% Pooling method comparisons
\textbf{Pooling method comparisons}. In Table~\ref{tab:pooling_imagenet_tests}, we compare multiple pooling methods across six networks. All networks were trained from scratch. SoftPool performs similarly to learnable approaches without requiring additional convolutions, while outperforming stochastic methods. These classification accuracies correlate with image similarities in Table~\ref{tab:times_ssim_psnr}. This enforces the notion that pooling methods that retain information will also improve classification accuracy.

\subsection{Multi-layer ablation study}
\label{sec:experiments::multi-layer}

In order to better understand how SoftPool affects the network performance at different depths, we use an InceptionV3 model \cite{szegedy2016rethinking} which integrates pooling in its layer structure. We systematically replace the max-pool operations within Inception blocks at different network layers.

% Per-layer gains
From the top-1 and top-5 results summarized in Table~\ref{tab:inception_tests_progressive}, we observe that the accuracy increases with the number of pooling layers that are replaced with SoftPool. An average increase of 0.23\% in top-1 accuracy is obtained with single layer replacements. The final top-1 accuracy surge between the original network with max-pool (N) and the SoftPool model (VII) is +1.59\%. This shows that SoftPool can be used as direct replacement regardless of the network depth.

\subsection{Classification performance on video data}
\label{sec:experiments::video}

% The challenge of downsampling in videos.
Finally, we demonstrate the merits of SoftPool in handling spatio-temporal data. Specifically, we address action recognition in videos where the input to the network is a stack of subsequent video frames. Representing time-based features stands as a major challenge in action recognition research \cite{stergiou2019analyzing}. The main challenge in space-time data downsampling is the inclusion of key temporal information without impacting the spatial quality of the input.

% Tests on 3D Convolutions
In this experiment, we use popular time-inclusive networks and replace the original pooling methods with SoftPool. Most space-time networks extend 2D convolutions to 3D to account for the temporal dimension. They use stacks of frames as inputs. For the tested networks with SoftPool, the only modification is that we use SoftPool to deal with the additional input dimension (see Section~\ref{sec:methodology::spacetime}).

% training
We trained most architectures from scratch on HACS \cite{zhao2019hacs} using the implementations provided by the authors. Results for Kinetics-700 and UCF-101 are fine-tuned from the HACS-trained models. We make exceptions for ir-CSN-101 and SlowFast, for which we used the networks trained for Kinetics-700 that were provided by the authors. ir-CSN-101 \cite{tran2019video} is pre-trained on IG65M \cite{ghadiyaram2019large}, a large dataset consisting of 65M video that is not publicly available. SlowFast \cite{feichtenhofer2019slowfast} is pre-trained on the full ImageNet dataset.

% Results of HACS
Results appear in Table~\ref{table:video_accuracies}. For three architectures, we report the performance on both the vanilla models and their counterparts with all pooling operations replaced by SoftPool. For r3d-50, a ResNet with 3D convolutions, using SoftPool increases the top-1 classification accuracy by 1.46\%. The accuracy performance also increases with 1.00\% and 1.63\% for the two SRTG models \cite{stergiou2021learn}. For the SRTG model with ResNet-(2+1) backbone, we achieve state-of-the-art performance on HACS. Also when using spatio-temporal data, SoftPool does not add computational complexity (GFLOPS), as demonstrated in Table~\ref{table:video_accuracies}.

% Results on Kinetics-700
For the performance on Kinetics-700, we observe important performance gains. An average of 1.22\% increase in top-1 accuracy is shown for the three models that have their pooling operations substituted by SoftPool. The best performing model is SRTG r3d-101 with SoftPool, which achieves a top-1 accuracy of 57.76\% and a top-5 accuracy score 77.84\%. These models also outperform state-of-the-art models such as SlowFast r3d-50 and ir-CSN-101.

When fine-tuning on UCF-101, the average accuracy gain is 0.66\% despite an almost saturated performance. SRTG r3d-101 with SoftPool is the best performing model with a top-1 accuracy of 98.06\% and top-5 of 99.82\%.

\section{Conclusions}
\label{sec:conclusions}

% Overview
We have introduced SoftPool, a novel pooling method that can better preserve informative features and, consequently, improves classification performance in CNNs. SoftPool uses the softmax of inputs within a kernel region where each of the activations has a proportional effect on the output. Activation gradients are relative to the weights assigned to them. Our operation is differentiable, which benefits efficient training. SoftPool does not require additional parameters nor increases the number of performed operations. We have shown the merits of the proposed approach through experimentation on image similarity tasks as well as on the classification of image and video datasets. We believe that the increase in classification performance combined with the low computation and memory requirements make SoftPool an excellent replacement for current pooling operations, including max and average pooling.

%\section*{Acknowledgments}
%\label{sec:acknowledgments}
%This publication is supported by the Netherlands Organization for Scientific Research (NWO) with a TOP-C2 grant for Automatic recognition of bodily interactions (ARBITER).

{\small
\bibliographystyle{ieee_fullname}
\bibliography{egbib}
}

% Pages are numbered in submission mode, and unnumbered in camera-ready
\ifcvprfinal\pagestyle{empty}\fi
\setcounter{section}{0}
\setcounter{equation}{0}
\setcounter{figure}{0}
\setcounter{table}{0}
\setcounter{page}{1}
\SupplementaryMaterials

\twocolumn[{%
\renewcommand\twocolumn[1][]{#1}%
\begin{center}
    \centering
    \textbf{\Large{Refining activation downsampling with SoftPool -- Supplementary material}}\\
    \hspace{2em}
    \includegraphics[width=\textwidth]{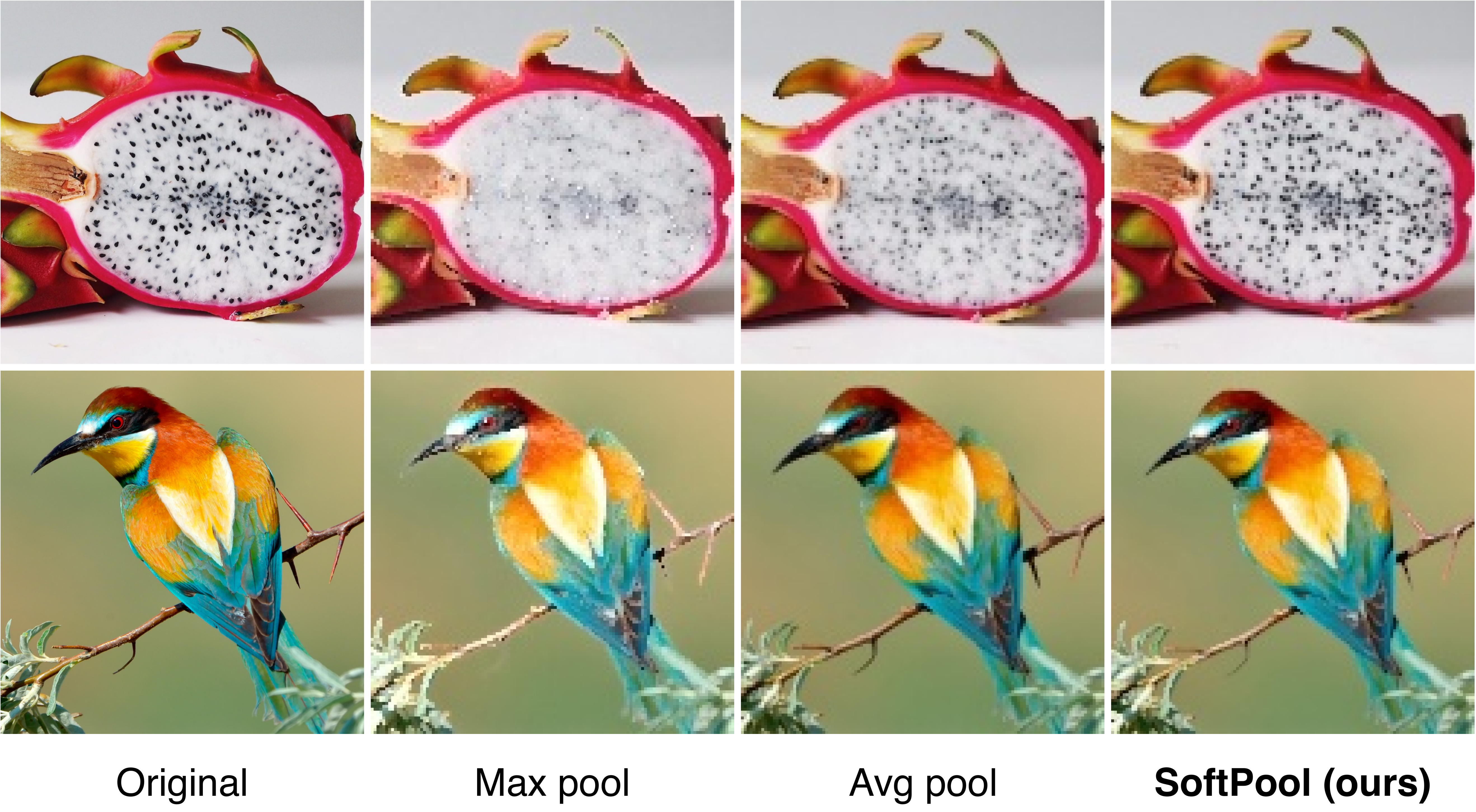}
    \captionof{figure}{\textbf{High resolution pooled images}. Original images are of size $1200 \times 1200$. The downsampled images were created with $\times 3$ pooling operations. To match the sizes of the original images and make the downsampling result more visible, we used inter area interpolation to resize the pooled images. This does not create a smoothing effect between neighboring pixels that are interpolated but rather populates new pixels based on the area relation.}
\label{fig:image_examples}
\end{center}%
}]

\section{Detail preservation}
\label{sec:detail}

We discussed and demonstrated the feature preservation capabilities of SoftPool in Section \textcolor{red}{3.3}. Here, we provide high-resolution images from examples in Figure \textcolor{red}{3} of the paper, in order to demonstrate clearer the effects of each pooling method. As it can be seen in Figure~\ref{fig:image_examples}, SoftPool can better capture detail in high-contrasting areas. Evidence of this can be seen in the top image where the dragon fruit seeds are not always preserved in the down-sampled images. For max pooling, most of the seeds are lost. This also applies to a certain extent to average pooling. By selecting the average within a region, features with high contrast are smoothed over, which reduces their effect significantly. In contrast, SoftPool preserves such regions in the sub-sampled outputs. By including part of the low-intensity regions in the output while weighting the high-intensity regions more, it can preserve the little-contrasting pixels. A similar pattern can also be seen with low-contrasting regions such as the bird's eye in the second row where, again, max pooling will highlight the high intensity features in the output while the downsampled output becomes less similar to that of the original image. Average pooling would instead make low-contrast features much more difficult to be recognized. SoftPool provides a balance between the two methods by weighting each part of the region respectfully to its intensity value.

\begin{figure}[ht]
    \centering
    \includegraphics[width=\linewidth]{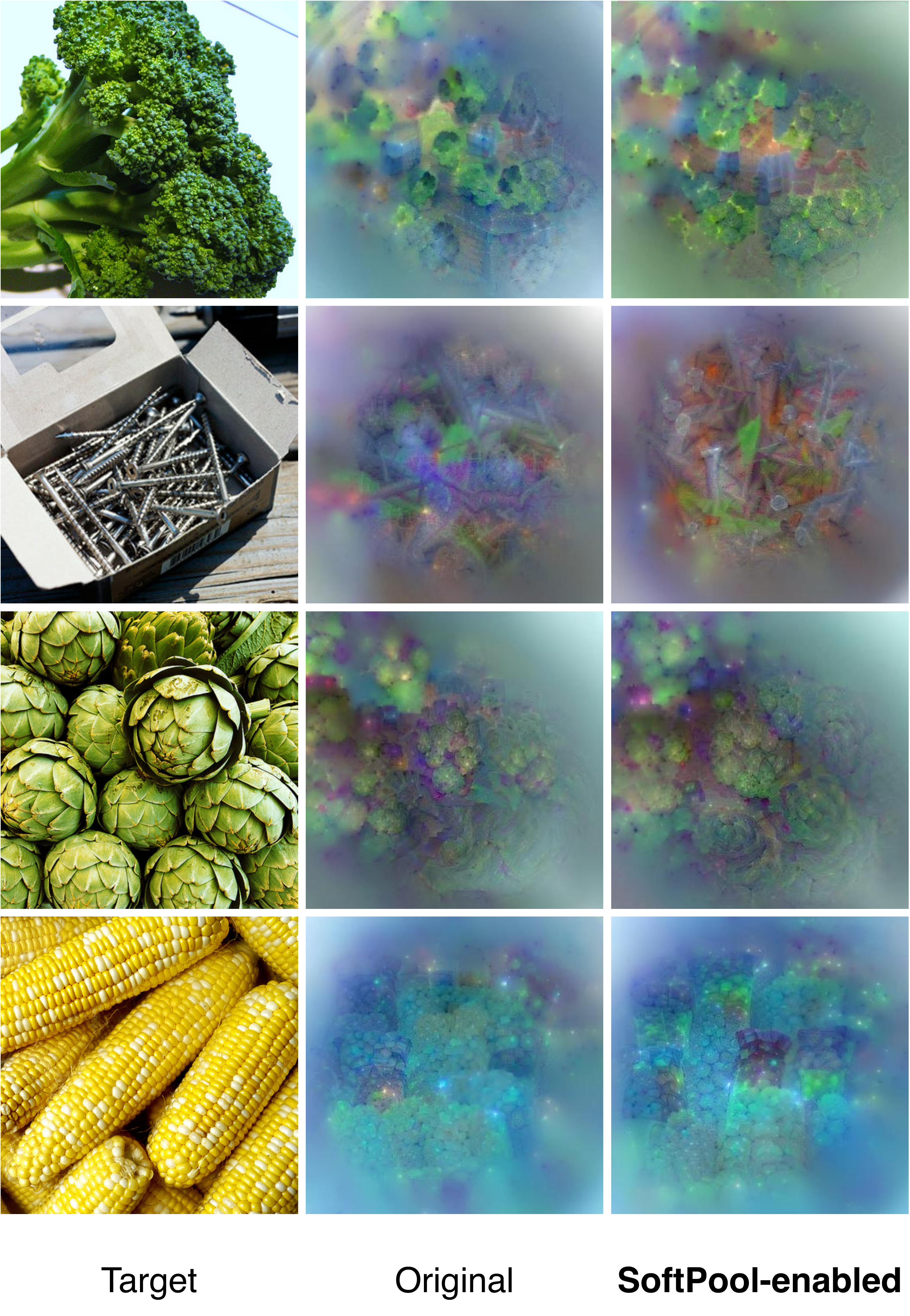}
    \caption{\textbf{Neuron activation maximization for InceptionV3 with original and SoftPool pooling layers.} We maximized the top-10 neurons \protect\citeS{erhan2009visualizing,stergiou2021mind} of the final block (\textit{Mixed7c}) in InceptionV3 \cite{szegedy2016rethinking}. Target top-10 neuron combinations for each row were selected for ImageNet1K classes ``broccoli'', ``nails'', ``artichoke'' and ``corn''. }
    \label{fig:neurone_opt}
\end{figure}

\section{Model feature visualization}
\label{sec:image_visualizatons}

% Activation maximization overview
Learned feature interpretability aims at understanding the features that networks associate with each class. One technique is \textit{activation maximization} \citeS{erhan2009visualizing} which creates a synthetic image by maximizing the activations relating to a specific neuron. Neurons could correspond to a specific class \citeS{simonyan2013deep} or features within the feature extractor \citeS{dosovitskiy2016inverting}.

% Training pipeline
To test the representation capabilities of networks with SoftPool, we use InceptionV3 \cite{szegedy2016rethinking} as a backbone model and visualize the top-10 most informative features. To train, we initialize an image with random noise, which we use as input for each of the two tested networks. During the training process, the image is optimized with the maximization of the activations of the top-10 kernels in the final InceptionV3 block (\textit{Mixed7c}) as objective function. The top-10 kernels are selected based on the highest average activations across all class examples. To eliminate additional noise, we use a mask-based approach similar to Wei \textit{et al.} \citeS{wei2015understanding}. However, in our setting, the mask is responsible for reducing the size of the gradient vectors for regions that are further away from the image center.

% Training setup
We used input images of size $512 \times 512$ in order to get higher-definition features. We use an SGD optimizer with an initial learning rate of 0.1 with a linear decrease to 0.01 over 2,000 total iterations. We use a weight decay of 1e-6. The weights for both models were initialized from those shown in Table \textcolor{red}{6} of the paper.

% Observations
We show the outputs in Figure~\ref{fig:neurone_opt} for four ImageNet1K classes: ``broccoli'', ``nails'', ``artichoke'' and ``corn''. We include in the left column the ImageNet1K images with the highest activations to provide representative examples. The majority of the features are fairly similar between the two models. Since SoftPool does not change the overall architecture nor the parameters, a high degree of similarity is expected. However, in cases such as the heads of the nails, we do notice a better definition of the objects. We also notice for the artichoke class that the structure of the petals and the thorns are more easy to distinguish from the network with SoftPool layers compared to the network with the original pooling layers. Although the differences remain small, by only changing the downsampling method used by the network, it can affect the robustness and improve the feature interpretability to a degree.

\section{Spatio-temporal volume pooling}
\label{sec:space_time_pool}

% Challenges in spatio-temporal pooling
Pooling operations in time-inclusive volumes (videos) face the additional challenge of encoding time in the output. One large problem is between-frame motion as it can significantly impact the representation of spatial features within frames in the sub-sampled volume.

% SoftPool in time-inclusive volumes
We show in Figure~\ref{fig:spatio_temporal_examples} with four different examples the effects of spatio-temporal pooling with average pooling, max-pool, and SoftPool operations. As none of the methods is tailored towards completely alleviating the effects of encoded motion in the pooled output, they are visible in edges and regions where cross-frame motion exists. However, differences between the three methods become apparent.

\begin{figure*}[!htb]
    \begin{minipage}[b]{0.15\textwidth}
        \bigbreak
        \bigbreak
        \bigbreak
        \bigbreak
        Parkour\\
        \bigbreak
        \bigbreak
        \bigbreak
        \bigbreak
        \bigbreak
        \bigbreak
        \bigbreak
        Running\\
        \bigbreak
        \bigbreak
        \bigbreak
        \bigbreak
        \bigbreak
        \bigbreak
        \bigbreak
        Dodgeball\\
        \bigbreak
        \bigbreak
        \bigbreak
        \bigbreak
        \bigbreak
        \bigbreak
        \bigbreak
        Table tennis\\
        \bigbreak
        \bigbreak
        
    \end{minipage}
    \hfill
    \setcounter{figure}{2}
    \begin{minipage}[t]{.2\textwidth}
    \centering
        \includegraphics[width=\textwidth]{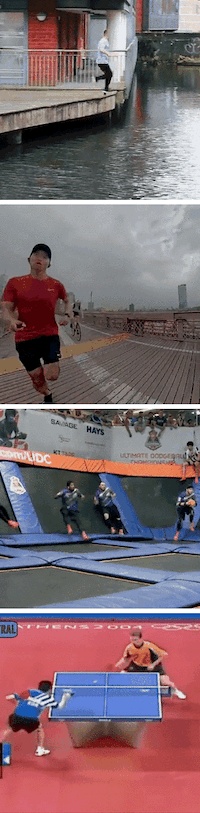}
        \captionsetup{labelformat=empty}
        \caption{a. Original}
        \label{fig:spatio_temporal_a}
    \end{minipage}
    \hfill
    \setcounter{figure}{2}
    \begin{minipage}[t]{.2\textwidth}
    \centering
        \includegraphics[width=\textwidth]{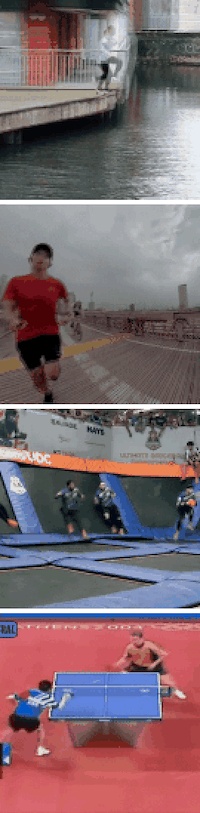}
        \captionsetup{labelformat=empty}
        \caption{b. Avg. pool (3D)}
        \label{fig:spatio_temporal_b}
    \end{minipage}
    \hfill
    \setcounter{figure}{2}
    \begin{minipage}[t]{.2\textwidth}
    \centering
        \includegraphics[width=\textwidth]{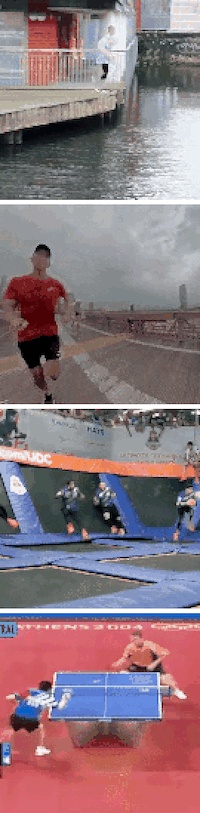}
        \captionsetup{labelformat=empty}
        \caption{c. Max-pool (3D)}
        \label{fig:spatio_temporal_c}
    \end{minipage}
    \hfill
    \setcounter{figure}{2}
    \begin{minipage}[t]{.2\textwidth}
    \centering
        \includegraphics[width=\textwidth]{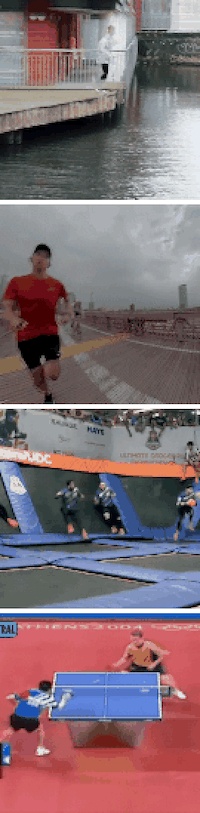}
        \captionsetup{labelformat=empty}
        \caption{d. SoftPool (3D)}
        \label{fig:spatio_temporal_d}
    \end{minipage}
    \setcounter{figure}{2}
    \begin{minipage}[t]{\linewidth}
    \vspace{.5 cm}
    \centering
        \includegraphics[width=\linewidth]{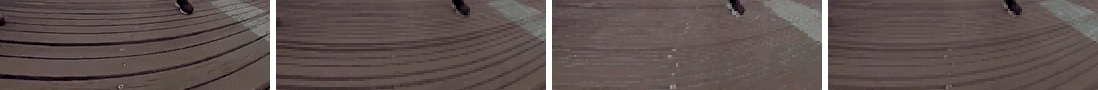}
        \includegraphics[width=\linewidth]{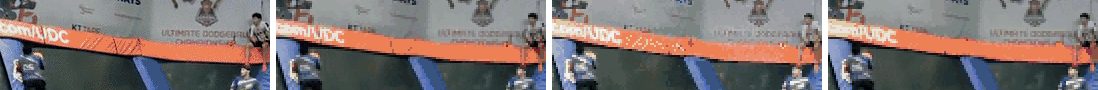}
        \captionsetup{labelformat=empty}
        \caption{e. Zoomed-in frame regions}
        \label{fig:spatio_temporal_e}
    \end{minipage} 
    \setcounter{figure}{2}
\caption{\textbf{Spatio-temporally downsampled videos.} Sub-sampling over spatio-temporal volumes compared to original videos (a), and downsampled with average pooling (b), maximum pooling (c) and our proposed SoftPool (d). Two zoomed-in frame regions appear in (e).% (The figure is best viewed on animation-enabled PDF viewers such as Adobe Acrobat).
}
\label{fig:spatio_temporal_examples}
\end{figure*}

% Zoomed-in differences
We demonstrate part of these differences in Figure~\ref{fig:spatio_temporal_e}(e). Where we include two cases of zoomed-in frame regions that show some variance based on the pooling method used. In the top one, gaps in the wooden planks of the floor are significantly less distinguishable within the max-pooled frame. Consequently, in the average pooled frame region, the nails are not visible at all anymore. This effect is in line with our observations for image-based downsampling of high-contrast and low-contrast regions. In contrast, SoftPool preserves features in both cases, which allows for the extraction of representative features after pooling.

\begin{figure*}[ht]
    \centering
    \includegraphics[width=.48\linewidth]{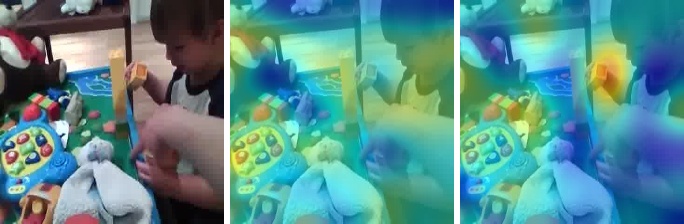}
    \includegraphics[width=.48\linewidth]{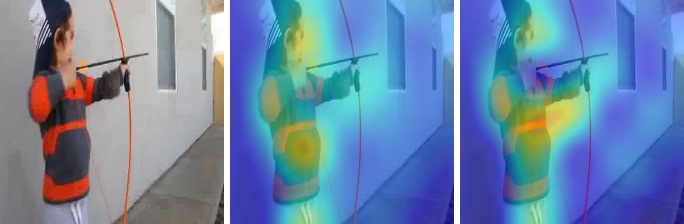}
    
    \includegraphics[width=.48\linewidth]{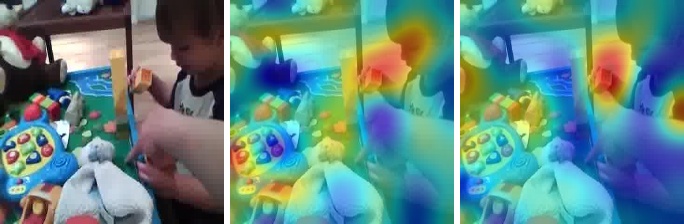}
    \includegraphics[width=.48\linewidth]{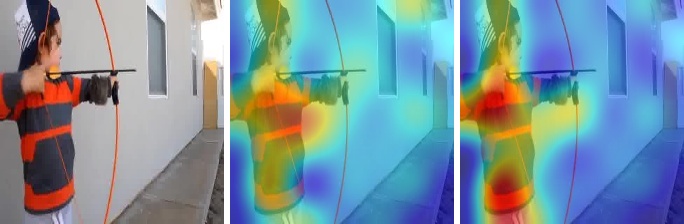}
    
    \includegraphics[width=.48\linewidth]{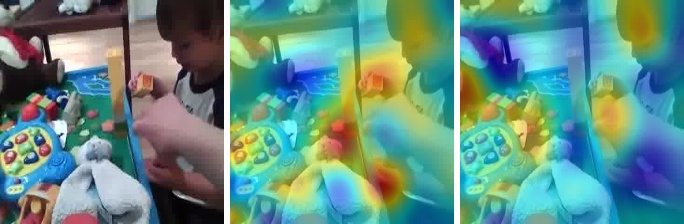}
    \includegraphics[width=.48\linewidth]{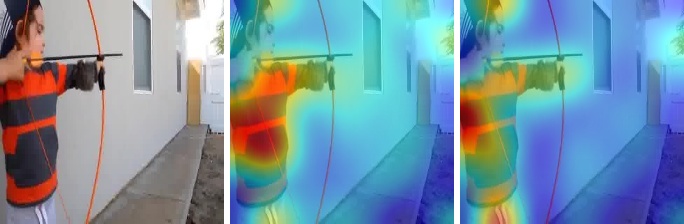}
    
    \includegraphics[width=.48\linewidth]{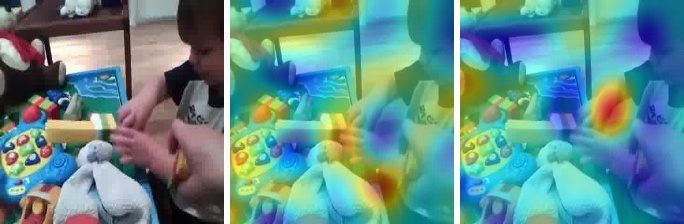}
    \includegraphics[width=.48\linewidth]{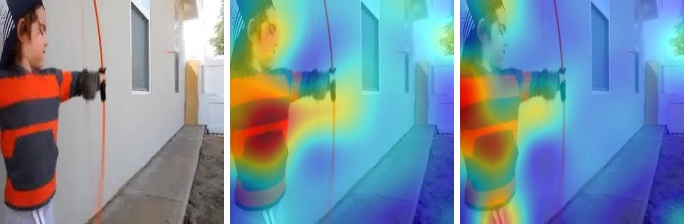}
    
    \includegraphics[width=.48\linewidth]{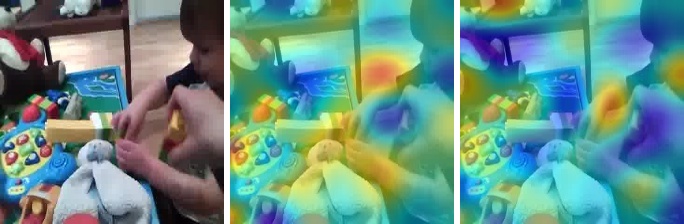}
    \includegraphics[width=.48\linewidth]{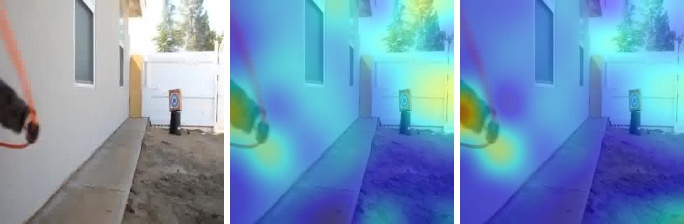}
    
    Source video \hspace{1 cm} original \hspace{1.2 cm} \textbf{with SoftPool} \hspace{1cm} Source video \hspace{1 cm} original \hspace{1.2 cm} \textbf{with SoftPool}
\caption{\textbf{Spatio-temporal saliency region visualizations for r3d-50 with and without SoftPool.} Class Feature Pyramids \protect\citeS{stergiou2019class,stergiou2019saliency} were used to generate the regional activations in the final \textit{conv} layer of r3d-50.% (The figure is best viewed on animation-enabled PDF viewers such as Adobe Acrobat).
\label{fig:saliency}}
\end{figure*}

\section{Time-inclusive salient regions}
\label{sec:space_time_saliency}

% Motivation for spatio-temporal saliency comparisons
To better understand the use of SoftPool in 3D-CNNs, we study the spatio-temporal regions that the network finds more informative. Similar to the activation maximization visualizations for images, we use a fixed network structure as a backbone and only study the variations produced by replacing the original pooling operations with SoftPool. For the visualizations in Figure~\ref{fig:saliency}, we use the r3d-50 network from Table \textcolor{red}{5} in the paper. The examples are sampled from the Kinetics-700 dataset from the classes ``building lego'' and ``archery''.

% Overview of results
Based on the examples presented in Figure~\ref{fig:saliency}, there are no significant differences in the salient regions. However, in multi-object scenes such as for the ``building lego'' class, the regional focus of the SoftPool network is shown to be a bit more distinct towards the region where there is a clear definition of the action performed (i.e. the hand with the lego brick). The case of ``archery'' exhibit small amounts of variations, with both either focusing on the main actor within the video.

\section{Embedding spaces visualizations}
\label{sec:embedding_space}

% Set-up
We provide t-SNE \citeS{maaten2008visualizing} visualizations of feature embeddings from an InceptionV3 model with original pooling operations and their counterparts with all pooling operations replaced by SoftPool. We use the averaged feature vectors of the final block in InceptionV3 (\textit{Mixed7c}) with a reduced dimensionality of 50 channels produced by PCA \citeS{jolliffe2003principal} and then perform t-SNE. We further perform k-means clustering \citeS{lloyd1982least} to better represent the different sub-clusters within the embedding space. In well-defined feature spaces, images in clusters that are closer together should be more similar.

% Resulting visualizations
The visualizations in Figures~\ref{fig:tsne_kmeans_a}-\ref{fig:tsne_kmeans_c} show distinct embeddings for the two networks. While structurally both model yield similar embeddings, for some classes the differences are more apparent. For example, class ``jack-o-lantern'' (Figure~\ref{fig:tsne_kmeans_b}) and ``sax'' (Figure~\ref{fig:tsne_kmeans_c}) show more compact representations when SoftPool is used.

%For the ``bald eagle'' class in Figure~\ref{fig:tsne_kmeans_a}, the vanilla InceptionV3 network does not distinguish between eagles that are flying (being the sky, wings spread) and eagles that are in rest (in a branch or on the ground). This is also true for the SoftPool network with the addition of a cluster that included multiple eagles flying rather than a singular eagle. An additional example is also seen in the  ``jack-o-latern'' class where in the SoftPool case the feature space is elongated with each of the cluster being fairly close to each other. The two ends of the ellipse are based on the overall image brightness with night and day images. This is also present for the original network as well however with a much less visible intersecting regions between clusters.  

\section{Error rates and statistical significance}
\label{sec:mcnemar}

% General McNemar test formulation
We further evaluate the statistical significance of the validation accuracy rates achieved in Table \textcolor{red}{2} in the context of the classification performance between the original models and models with pooling layers replaced by SoftPool. We perform a McNemar's test \cite{edwards1948note,mcnemar1947note}, which is based on a null hypothesis ($H_{0}$) corresponding to accuracy homogeneity between the two models. It is calculated based on a contingency table holding the number of correct or incorrect class prediction instances with respect to each model. The tests studies the number of disagreements in the predictions between the two models. The Chi-Square statistic ($\chi^{2}$) is then calculated based on the number of model $i$ correct predictions against model $j$ incorrect predictions ($n_{ij}$) and model $j$ correct against model $i$ incorrect predictions ($n_{ji}$). This is expressed as:

\begin{equation}
    \chi^{2} = \frac{(|n_{ij}-n_{ji}|-1)^{2}}{n_{ij}+n_{ji}}
\end{equation}

We note that the null hypothesis ($H_{0}$) can be rejected with different significance levels based on the Chi-Square distribution table \citeS{chidisttable}. As the statistic is calculated with a single degree of freedom, $\chi^{2}$ values of $\{ 3.84, 5.02, 6.63 \}$ correspond to equivalent probabilities of $\{ 95\%, 97.5\%, 99.0\% \}$ that the two methods indeed differ.

\begin{table*}[htp]
    \subfloat[ResNet18]{
        \centering
        \begin{tabular}{l l p{3.5em} | p{3.5em} p{3em}}
        & & \multicolumn{2}{c}{Original} & \multirow{2}{*}{Total}\\[.15em]
        & & Correct & Incorrect & \\[.15em] \cline{3-5}
        \multirow{2}{*}{SoftPool} & \multicolumn{1}{c|}{Correct} & \multicolumn{1}{c|}{31838} & \multicolumn{1}{c|}{3795} & \multicolumn{1}{c|}{35633}\\ [.15em] \cline{2-5}
         & \multicolumn{1}{c|}{Incorrect} & \multicolumn{1}{c|}{3011} & \multicolumn{1}{c|}{11356} & \multicolumn{1}{c|}{14367}\\ [.15em] \cline{3-5}
        \multicolumn{2}{c|}{Total} & \multicolumn{1}{c|}{34849} & \multicolumn{1}{c|}{15151} & \multicolumn{1}{c}{50000} \\ [.15em] \cline{3-4}
       \end{tabular}
       \label{tab:mcnemar_resnet18}
    }
    \hfill
    \subfloat[ResNet34]{
        \centering
        \begin{tabular}{l l p{3.5em} | p{3.5em} p{3em}}
        & & \multicolumn{2}{c}{Original} & \multirow{2}{*}{Total}\\[.15em]
        & & Correct & Incorrect & \\[.15em] \cline{3-5}
        \multirow{2}{*}{SoftPool} & \multicolumn{1}{c|}{Correct} & \multicolumn{1}{c|}{34090} & \multicolumn{1}{c|}{3246} & \multicolumn{1}{c|}{37336}\\ [.15em] \cline{2-5}
         & \multicolumn{1}{c|}{Incorrect} & \multicolumn{1}{c|}{2940} & \multicolumn{1}{c|}{9724} & \multicolumn{1}{c|}{12664}\\ [.15em] \cline{3-5}
        \multicolumn{2}{c|}{Total} & \multicolumn{1}{c|}{37030} & \multicolumn{1}{c|}{12970} & \multicolumn{1}{c}{50000} \\ [.15em] \cline{3-4}
       \end{tabular}
       \label{tab:mcnemar_resnet34}
    }
    \\
    \subfloat[ResNet50]{
        \centering
        \begin{tabular}{l l p{3.5em} | p{3.5em} p{3em}}
        & & \multicolumn{2}{c}{Original} & \multirow{2}{*}{Total}\\[.15em]
        & & Correct & Incorrect & \\[.15em] \cline{3-5}
        \multirow{2}{*}{SoftPool} & \multicolumn{1}{c|}{Correct} & \multicolumn{1}{c|}{35472} & \multicolumn{1}{c|}{3203} & \multicolumn{1}{c|}{38675}\\ [.15em] \cline{2-5}
         & \multicolumn{1}{c|}{Incorrect} & \multicolumn{1}{c|}{2603} & \multicolumn{1}{c|}{8722} & \multicolumn{1}{c|}{11325}\\ [.15em] \cline{3-5}
        \multicolumn{2}{c|}{Total} & \multicolumn{1}{c|}{38075} & \multicolumn{1}{c|}{11925} & \multicolumn{1}{c}{50000} \\ [.15em] \cline{3-4}
       \end{tabular}
       \label{tab:mcnemar_resnet50}
    }
    \hfill
    \subfloat[ResNet101]{
        \centering
        \begin{tabular}{l l p{3.5em} | p{3.5em} p{3em}}
        & & \multicolumn{2}{c}{Original} & \multirow{2}{*}{Total}\\[.15em]
        & & Correct & Incorrect & \\[.15em] \cline{3-5}
        \multirow{2}{*}{SoftPool} & \multicolumn{1}{c|}{Correct} & \multicolumn{1}{c|}{36553} & \multicolumn{1}{c|}{2609} & \multicolumn{1}{c|}{39162}\\ [.15em] \cline{2-5}
         & \multicolumn{1}{c|}{Incorrect} & \multicolumn{1}{c|}{2272} & \multicolumn{1}{c|}{8566} & \multicolumn{1}{c|}{10838}\\ [.15em] \cline{3-5}
        \multicolumn{2}{c|}{Total} & \multicolumn{1}{c|}{38825} & \multicolumn{1}{c|}{11175} & \multicolumn{1}{c}{50000} \\ [.15em] \cline{3-4}
       \end{tabular}
       \label{tab:mcnemar_resnet101}
    }
    \\
    \subfloat[ResNet152]{
        \centering
        \begin{tabular}{l l p{3.5em} | p{3.5em} p{3em}}
        & & \multicolumn{2}{c}{Original} & \multirow{2}{*}{Total}\\[.15em]
        & & Correct & Incorrect & \\[.15em] \cline{3-5}
        \multirow{2}{*}{SoftPool} & \multicolumn{1}{c|}{Correct} & \multicolumn{1}{c|}{37247} & \multicolumn{1}{c|}{2373} & \multicolumn{1}{c|}{39620}\\ [.15em] \cline{2-5}
         & \multicolumn{1}{c|}{Incorrect} & \multicolumn{1}{c|}{1908} & \multicolumn{1}{c|}{8472} & \multicolumn{1}{c|}{10380}\\ [.15em] \cline{3-5}
        \multicolumn{2}{c|}{Total} & \multicolumn{1}{c|}{39155} & \multicolumn{1}{c|}{10845} & \multicolumn{1}{c}{50000} \\ [.15em] \cline{3-4}
       \end{tabular}
       \label{tab:mcnemar_resnet152}
    }
    \hfill
    \subfloat[DenseNet121]{
        \centering
        \begin{tabular}{l l p{3.5em} | p{3.5em} p{3em}}
        & & \multicolumn{2}{c}{Original} & \multirow{2}{*}{Total}\\[.15em]
        & & Correct & Incorrect & \\[.15em] \cline{3-5}
        \multirow{2}{*}{SoftPool} & \multicolumn{1}{c|}{Correct} & \multicolumn{1}{c|}{34097} & \multicolumn{1}{c|}{3836} & \multicolumn{1}{c|}{37933}\\ [.15em] \cline{2-5}
         & \multicolumn{1}{c|}{Incorrect} & \multicolumn{1}{c|}{2253} & \multicolumn{1}{c|}{9814} & \multicolumn{1}{c|}{12067}\\ [.15em] \cline{3-5}
        \multicolumn{2}{c|}{Total} & \multicolumn{1}{c|}{36350} & \multicolumn{1}{c|}{13650} & \multicolumn{1}{c}{50000} \\ [.15em] \cline{3-4}
       \end{tabular}
       \label{tab:mcnemar_densenet121}
    }
    \\
    \subfloat[DenseNet161]{
        \centering
        \begin{tabular}{l l p{3.5em} | p{3.5em} p{3em}}
        & & \multicolumn{2}{c}{Original} & \multirow{2}{*}{Total}\\[.15em]
        & & Correct & Incorrect & \\[.15em] \cline{3-5}
        \multirow{2}{*}{SoftPool} & \multicolumn{1}{c|}{Correct} & \multicolumn{1}{c|}{35160} & \multicolumn{1}{c|}{4237} & \multicolumn{1}{c|}{39397}\\ [.15em] \cline{2-5}
         & \multicolumn{1}{c|}{Incorrect} & \multicolumn{1}{c|}{3631} & \multicolumn{1}{c|}{6972} & \multicolumn{1}{c|}{10603}\\ [.15em] \cline{3-5}
        \multicolumn{2}{c|}{Total} & \multicolumn{1}{c|}{38791} & \multicolumn{1}{c|}{11209} & \multicolumn{1}{c}{50000} \\ [.15em] \cline{3-4}
       \end{tabular}
       \label{tab:mcnemar_densenet161}
    }
    \hfill
    \subfloat[DenseNet169]{
        \centering
        \begin{tabular}{l l p{3.5em} | p{3.5em} p{3em}}
        & & \multicolumn{2}{c}{Original} & \multirow{2}{*}{Total}\\[.15em]
        & & Correct & Incorrect & \\[.15em] \cline{3-5}
        \multirow{2}{*}{SoftPool} & \multicolumn{1}{c|}{Correct} & \multicolumn{1}{c|}{35074} & \multicolumn{1}{c|}{3407} & \multicolumn{1}{c|}{38481}\\ [.15em] \cline{2-5}
         & \multicolumn{1}{c|}{Incorrect} & \multicolumn{1}{c|}{2988} & \multicolumn{1}{c|}{8531} & \multicolumn{1}{c|}{11519}\\ [.15em] \cline{3-5}
        \multicolumn{2}{c|}{Total} & \multicolumn{1}{c|}{38062} & \multicolumn{1}{c|}{11938} & \multicolumn{1}{c}{50000} \\ [.15em] \cline{3-4}
       \end{tabular}
       \label{tab:mcnemar_densenet169}
    }
    \\
    \subfloat[ResNeXt50]{
        \centering
        \begin{tabular}{l l p{3.5em} | p{3.5em} p{3em}}
        & & \multicolumn{2}{c}{Original} & \multirow{2}{*}{Total}\\[.15em]
        & & Correct & Incorrect & \\[.15em] \cline{3-5}
        \multirow{2}{*}{SoftPool} & \multicolumn{1}{c|}{Correct} & \multicolumn{1}{c|}{36628} & \multicolumn{1}{c|}{2618} & \multicolumn{1}{c|}{39246}\\ [.15em] \cline{2-5}
         & \multicolumn{1}{c|}{Incorrect} & \multicolumn{1}{c|}{2153} & \multicolumn{1}{c|}{8601} & \multicolumn{1}{c|}{10754}\\ [.15em] \cline{3-5}
        \multicolumn{2}{c|}{Total} & \multicolumn{1}{c|}{38781} & \multicolumn{1}{c|}{11219} & \multicolumn{1}{c}{50000} \\ [.15em] \cline{3-4}
       \end{tabular}
       \label{tab:mcnemar_resnext50}
    }
    \hfill
    \subfloat[ResNeXt101]{
        \centering
        \begin{tabular}{l l p{3.5em} | p{3.5em} p{3em}}
        & & \multicolumn{2}{c}{Original} & \multirow{2}{*}{Total}\\[.15em]
        & & Correct & Incorrect & \\[.15em] \cline{3-5}
        \multirow{2}{*}{SoftPool} & \multicolumn{1}{c|}{Correct} & \multicolumn{1}{c|}{38633} & \multicolumn{1}{c|}{2587} & \multicolumn{1}{c|}{41220}\\ [.15em] \cline{2-5}
         & \multicolumn{1}{c|}{Incorrect} & \multicolumn{1}{c|}{1005} & \multicolumn{1}{c|}{7775} & \multicolumn{1}{c|}{8780}\\ [.15em] \cline{3-5}
        \multicolumn{2}{c|}{Total} & \multicolumn{1}{c|}{39638} & \multicolumn{1}{c|}{10362} & \multicolumn{1}{c}{50000} \\ [.15em] \cline{3-4}
       \end{tabular}
       \label{tab:mcnemar_resnext101}
    }
    \\
    \subfloat[wide-ResNet50]{
        \centering
        \begin{tabular}{l l p{3.5em} | p{3.5em} p{3em}}
        & & \multicolumn{2}{c}{Original} & \multirow{2}{*}{Total}\\[.15em]
        & & Correct & Incorrect & \\[.15em] \cline{3-5}
        \multirow{2}{*}{SoftPool} & \multicolumn{1}{c|}{Correct} & \multicolumn{1}{c|}{36638} & \multicolumn{1}{c|}{3113} & \multicolumn{1}{c|}{39751}\\ [.15em] \cline{2-5}
         & \multicolumn{1}{c|}{Incorrect} & \multicolumn{1}{c|}{2636} & \multicolumn{1}{c|}{7613} & \multicolumn{1}{c|}{10249}\\ [.15em] \cline{3-5}
        \multicolumn{2}{c|}{Total} & \multicolumn{1}{c|}{38781} & \multicolumn{1}{c|}{10726} & \multicolumn{1}{c}{50000} \\ [.15em] \cline{3-4}
       \end{tabular}
       \label{tab:mcnemar_wideresnet50}
    }
    \hfill
    \subfloat[InceptionV1]{
        \centering
        \begin{tabular}{l l p{3.5em} | p{3.5em} p{3em}}
        & & \multicolumn{2}{c}{Original} & \multirow{2}{*}{Total}\\[.15em]
        & & Correct & Incorrect & \\[.15em] \cline{3-5}
        \multirow{2}{*}{SoftPool} & \multicolumn{1}{c|}{Correct} & \multicolumn{1}{c|}{31826} & \multicolumn{1}{c|}{3897} & \multicolumn{1}{c|}{35723}\\ [.15em] \cline{2-5}
         & \multicolumn{1}{c|}{Incorrect} & \multicolumn{1}{c|}{3037} & \multicolumn{1}{c|}{11240} & \multicolumn{1}{c|}{14277}\\ [.15em] \cline{3-5}
        \multicolumn{2}{c|}{Total} & \multicolumn{1}{c|}{34863} & \multicolumn{1}{c|}{15137} & \multicolumn{1}{c}{50000} \\ [.15em] \cline{3-4}
       \end{tabular}
       \label{tab:mcnemar_inceptionv1}
    }
    \\
    \subfloat[InceptionV3]{
        \centering
        \begin{tabular}{l l p{3.5em} | p{3.5em} p{3em}}
        & & \multicolumn{2}{c}{Original} & \multirow{2}{*}{Total}\\[.15em]
        & & Correct & Incorrect & \\[.15em] \cline{3-5}
        \multirow{2}{*}{SoftPool} & \multicolumn{1}{c|}{Correct} & \multicolumn{1}{c|}{36341} & \multicolumn{1}{c|}{3029} & \multicolumn{1}{c|}{14874}\\ [.15em] \cline{2-5}
         & \multicolumn{1}{c|}{Incorrect} & \multicolumn{1}{c|}{2370} & \multicolumn{1}{c|}{8260} & \multicolumn{1}{c|}{10630}\\ [.15em] \cline{3-5}
        \multicolumn{2}{c|}{Total} & \multicolumn{1}{c|}{38711} & \multicolumn{1}{c|}{11289} & \multicolumn{1}{c}{50000} \\ [.15em] \cline{3-4}
       \end{tabular}
       \label{tab:mcnemar_inceptionv3}
    }
     \caption{Error summaries of original models and models with pooling replaced with SoftPool.}
     \label{tab:mcnemar}
\end{table*}

% Table explanation
Prediction distributions of each model pair are presented in Table~\ref{tab:mcnemar} and the resulting $\chi^{2}$ statistics and $\rho$ homogeneity probabilities are presented in Table~\textcolor{red}{4} in the main text. Based on the very low homogeneity probability $\ll 0.01\%$ \citeS{craparo2007significance,fisher1992statistical,neyman1937outline}, the differences between the original networks and the networks that have been re-trained with SoftPool cannot be attributed to statistical errors.

\section{Implementation Details}
\label{sec:implementation}

\textbf{Range definitions}. The exponential weighting of activations can correspond to the produced values being smaller than the type's (16, 32, 64-bit) precision level lower threshold. This can either result in a computational underflow or in a zero-valued dividend. For this reason, we include additional checks that each produced exponentially scaled activation $e^{\textbf{a}_{i}}$ and their resulting weight mask $\textbf{w}_{i}$, based on which their values ($x$) are transformed, to $x = max(0,x)$. The sum of the weights is constrained similarly, based on $x = max(x_{min},x)$, where $x_{min}$ is the lowest limit based on type chosen to ensure non-zero dividend. We note that these checks do need to address changes in the activation functions used. The current tested networks use ReLU activations which have lower bounds of zero. When considering other non-zero or negative-valued lower bound functions, the transformations need to be adjusted accordingly.

\textbf{Computational description}. As our implementation is native to CUDA-enabled devices, we are able to achieve inference times close to those of native methods such as average and maximum pooling. However, the parallelization capabilities of SoftPool allow for running times similar to those of average pooling with $O(1)$. This is based on the fact that operations can be performed through a matrix over the kernel region. This is beneficial for processes that have parallelization as a backbone (CUDA). In contrast, max pooling has complexity of at least $O(n)$, as the selection of the maximum value within the region can only be performed through the sequential consideration of each pixel within the region. 

\textbf{Testing environment}. All of our tests were done with half floating point precision (float16) instead of single-point (float32) for better memory utilization during the model training phase. The batch sizes are split equally with 64 images per GPU. Our testing environment consists of an AMD Threadripper 2950X with 2400MHz RAM frequency and four Nvidia 2080 TIs.

{\small
\bibliographystyleS{plain}
\bibliographyS{egbib}
}

\begin{figure*}[!htb]
    \begin{minipage}[t]{.5\textwidth}
    \centering
        \includegraphics[trim=20px 20px 20px 20px, clip,width=\textwidth]{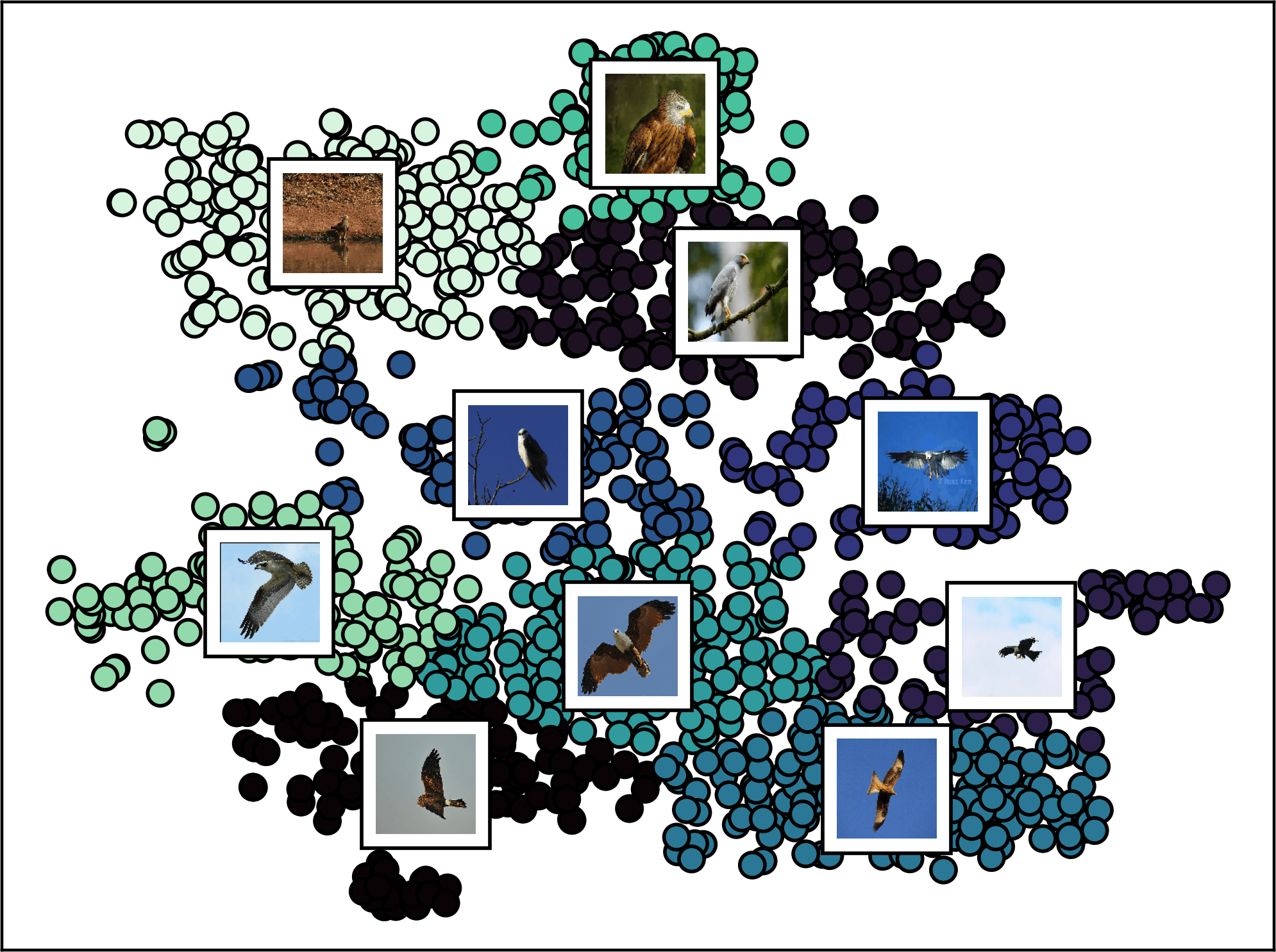}
        \includegraphics[trim=20px 20px 20px 20px, clip,width=\textwidth]{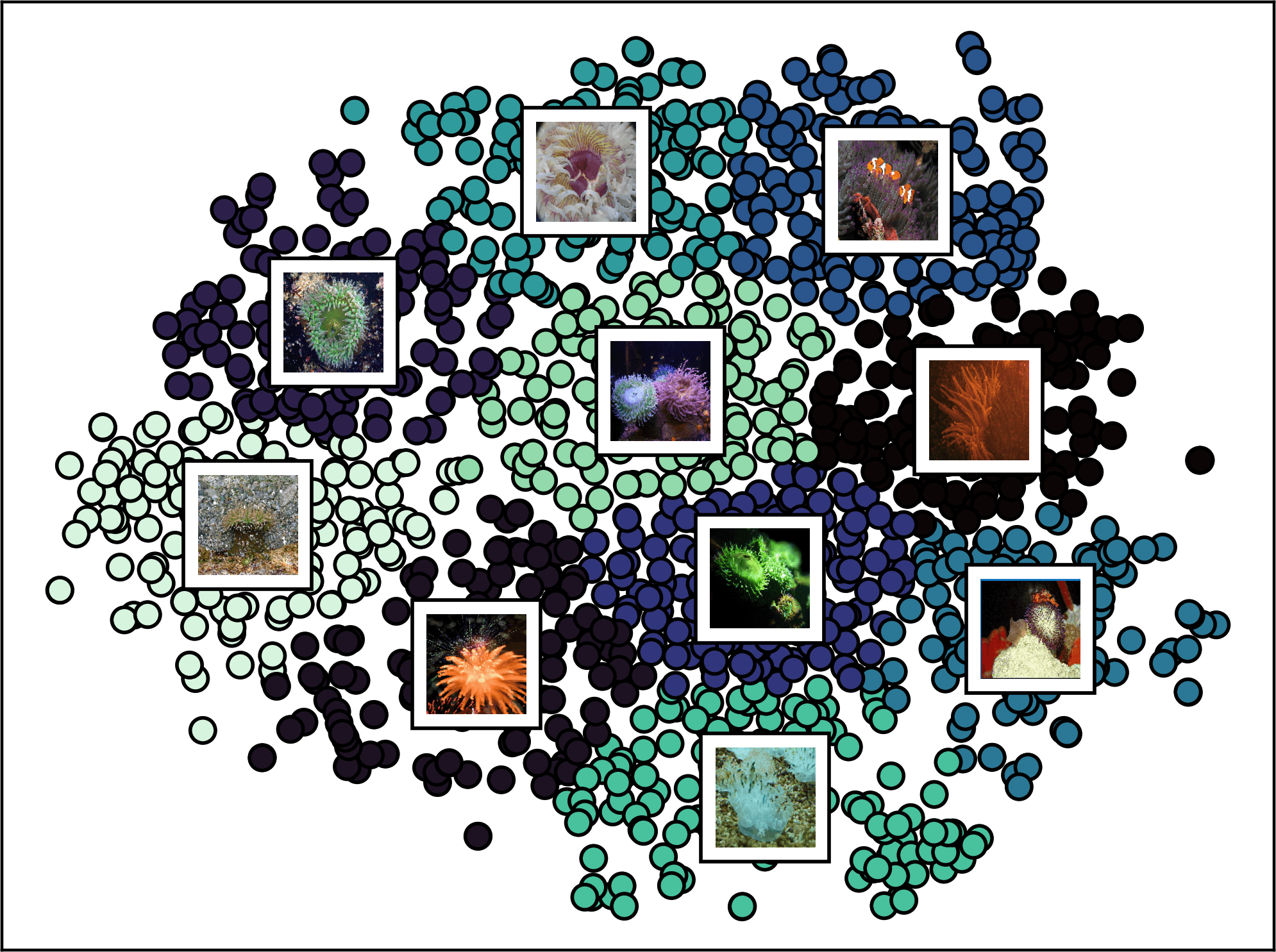}
        \includegraphics[trim=20px 20px 20px 20px, clip,width=\textwidth]{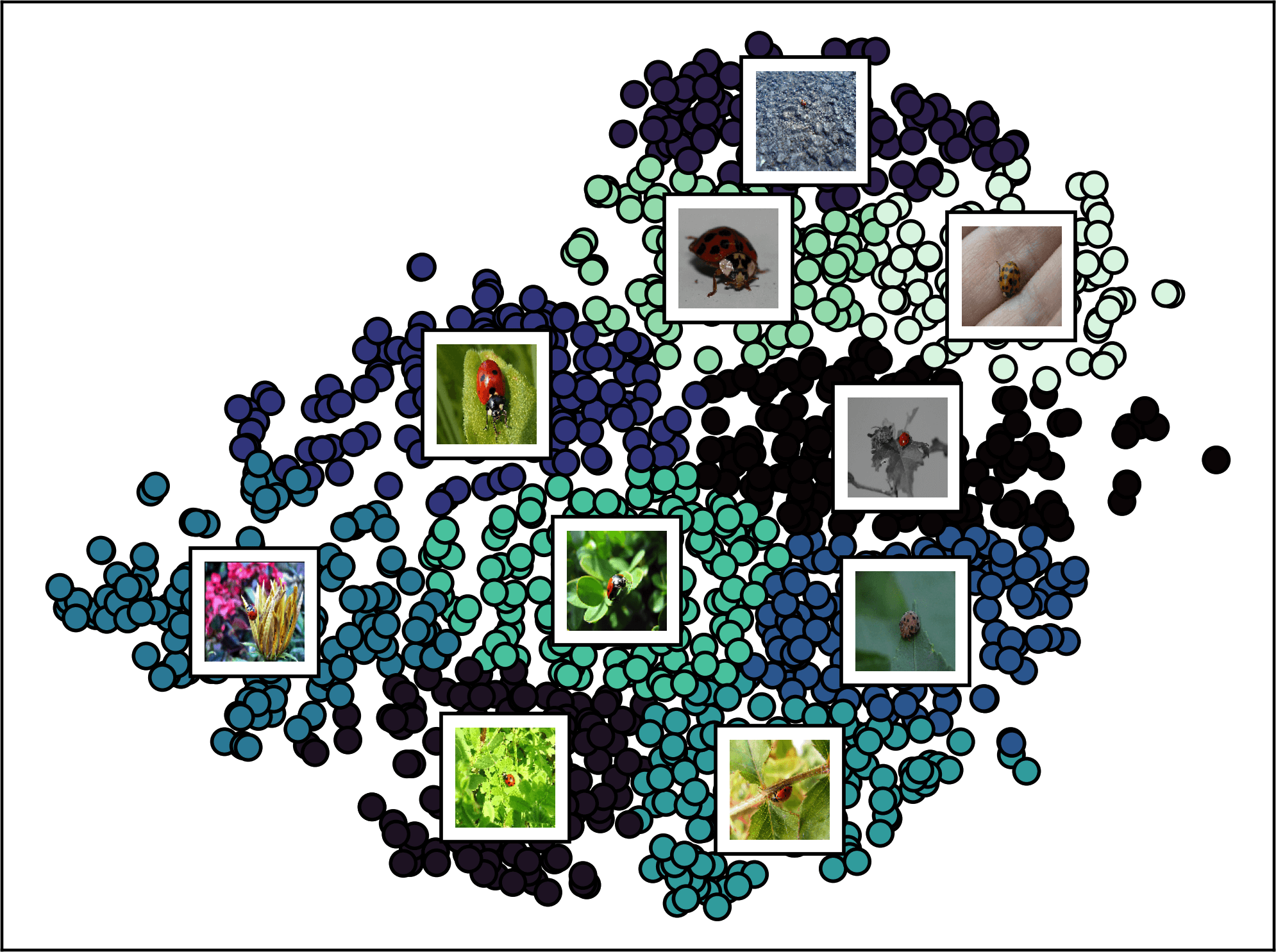}
        \captionsetup{labelformat=empty}
        \caption{a. Original}
        \label{fig:tsne_original_a}
    \end{minipage}
    \hfill
    \begin{minipage}[t]{.5\textwidth}
    \centering
        \includegraphics[trim=20px 20px 20px 20px, clip,width=\textwidth]{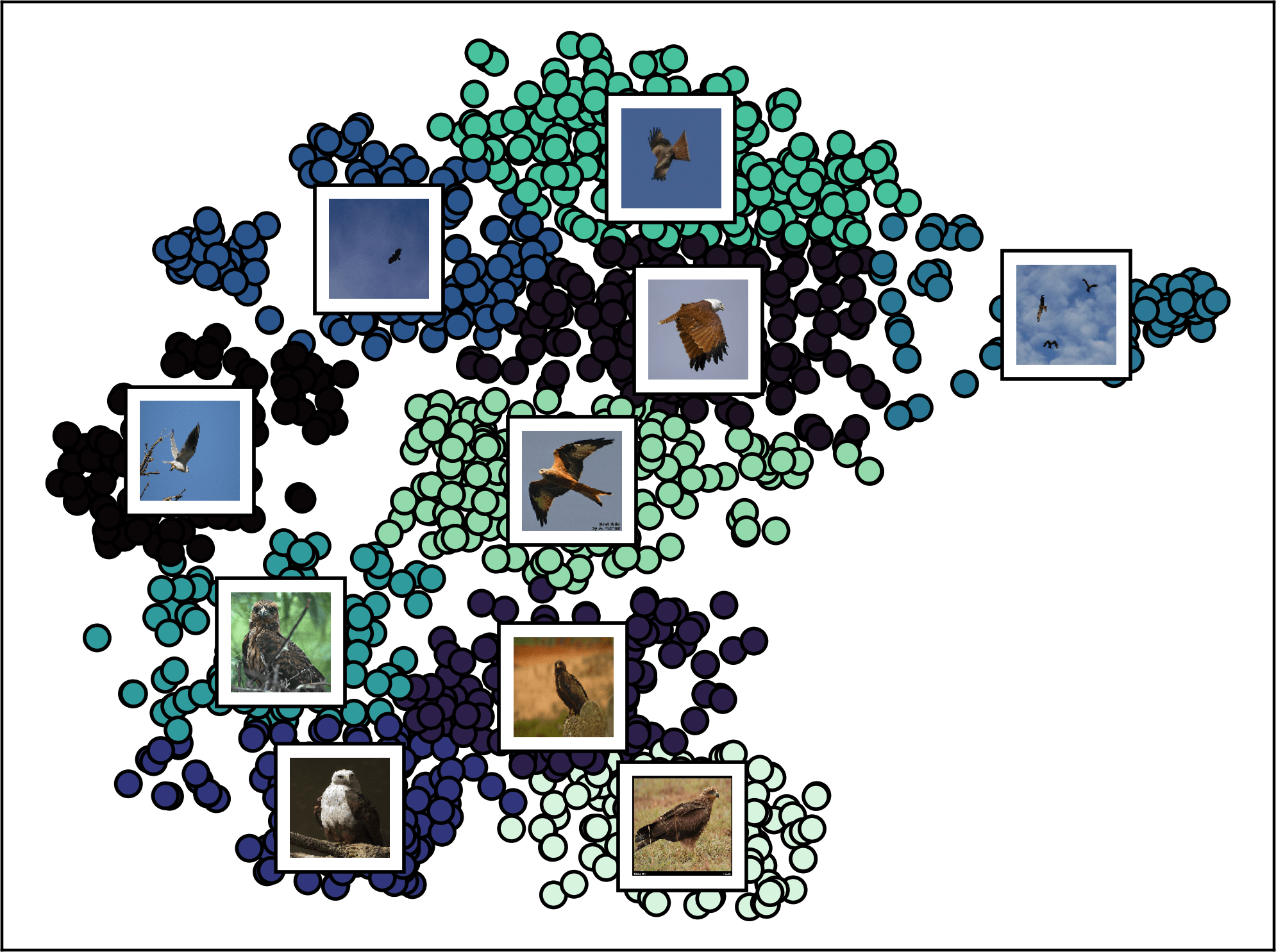}
        \includegraphics[trim=20px 20px 20px 20px, clip,width=\textwidth]{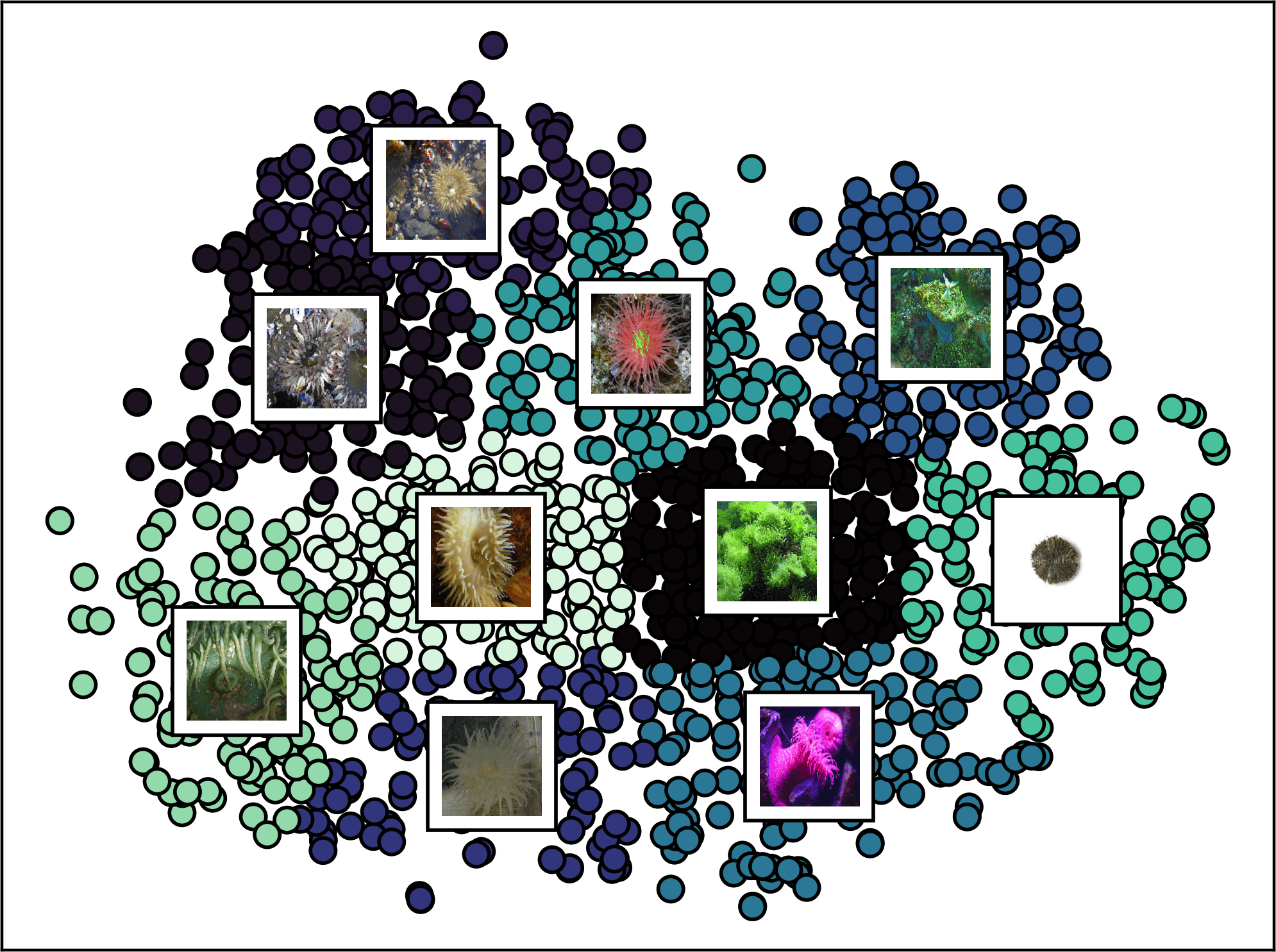}
        \includegraphics[trim=20px 20px 20px 20px, clip,width=\textwidth]{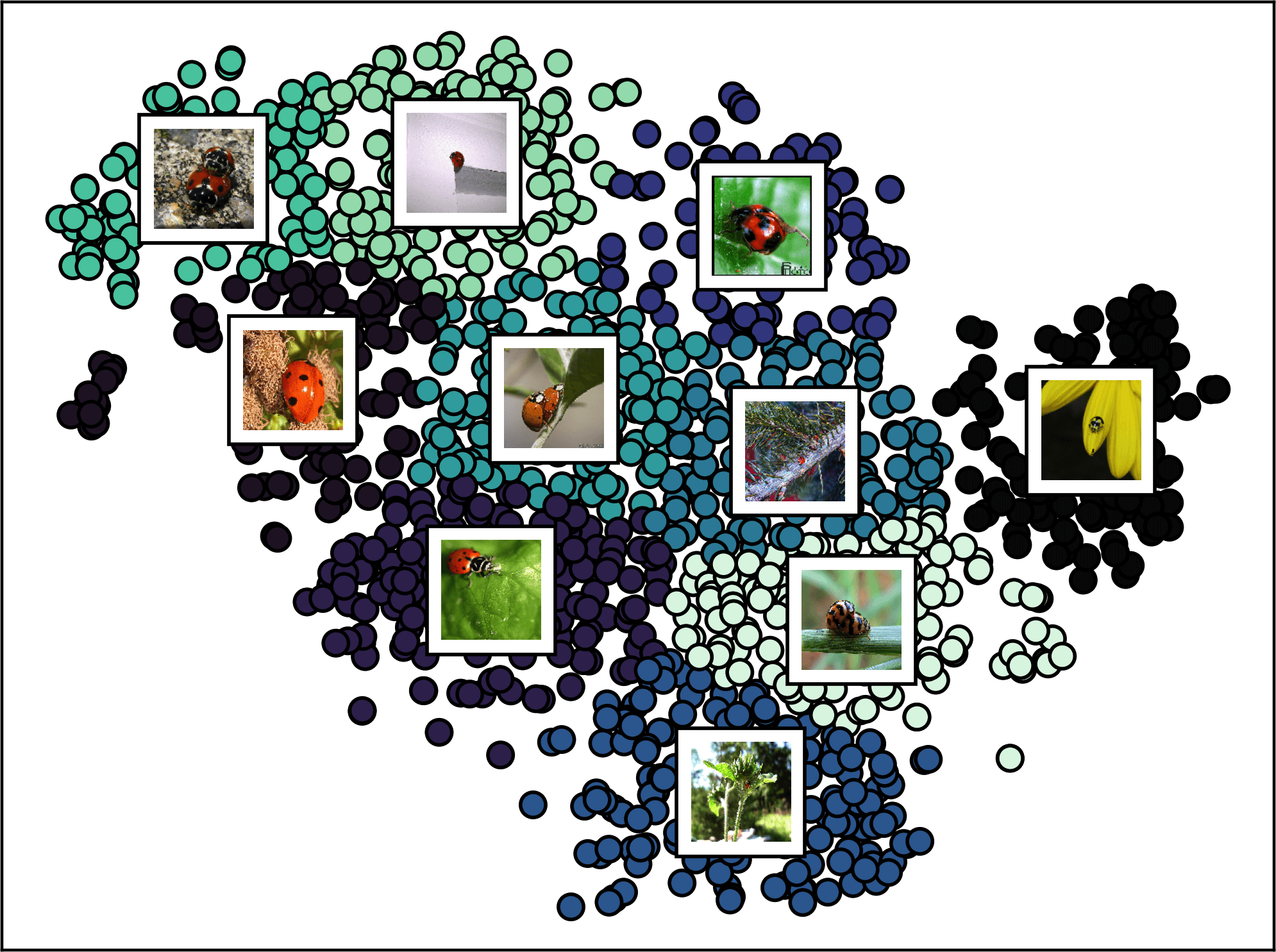}
        \captionsetup{labelformat=empty}
        \caption{b. SoftPool}
        \label{fig:tsne_softpool_a}
    \end{minipage}
\setcounter{figure}{4}
\caption{\textbf{t-SNE feature embeddings for InceptionV3 with and without SoftPool}. ImageNet1K classes ``bald eagle'', ``sea anemone'' and ``ladybug''. Cluster centers are found with k-means to better visualize the feature space. Images displayed are the closest examples for each cluster center.}
\label{fig:tsne_kmeans_a}
\end{figure*}

\begin{figure*}[!htb]
    \begin{minipage}[t]{.5\textwidth}
    \centering
        \includegraphics[trim=20px 20px 20px 20px, clip,width=\textwidth]{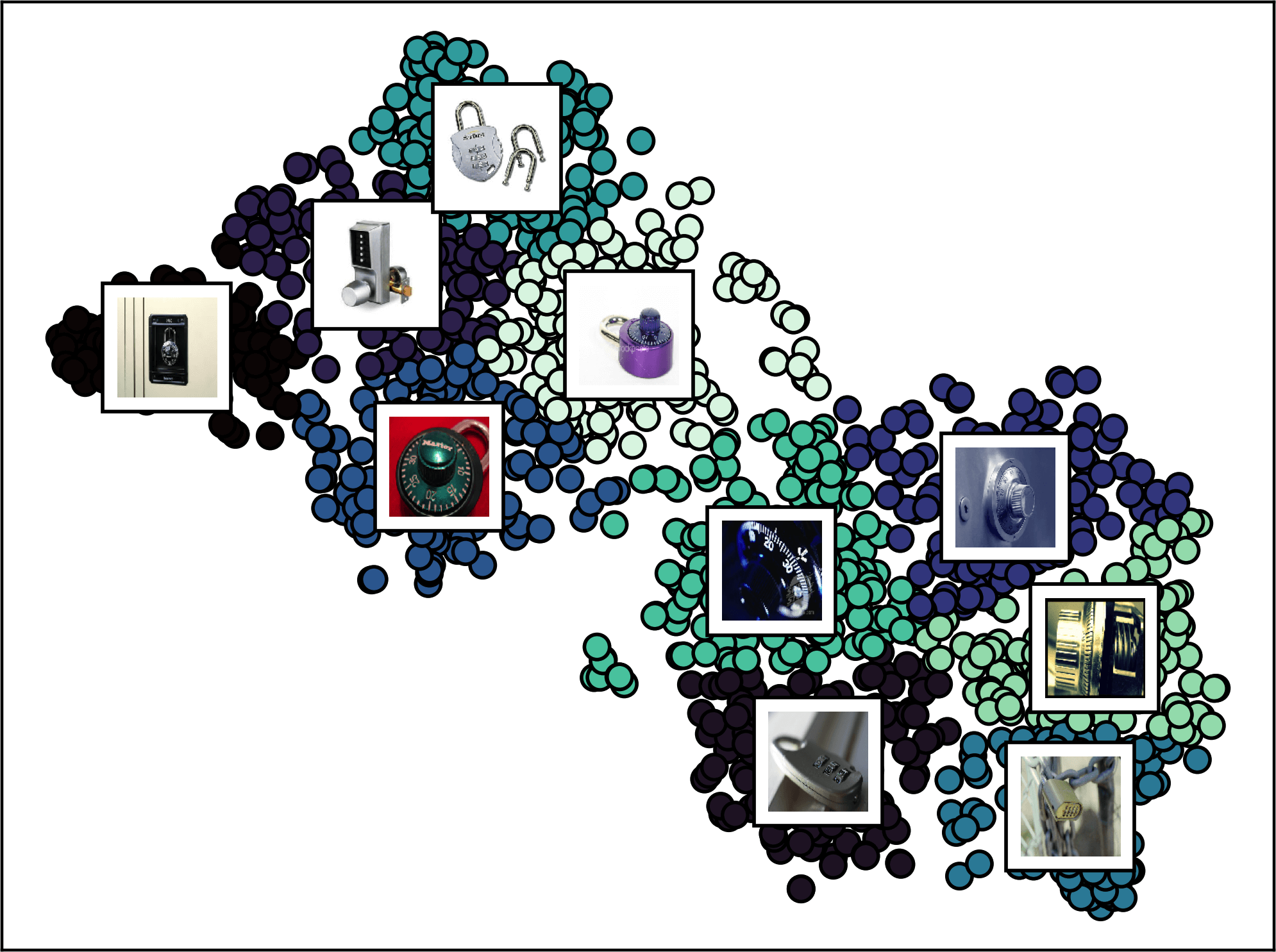}
        \includegraphics[trim=20px 20px 20px 20px, clip,width=\textwidth]{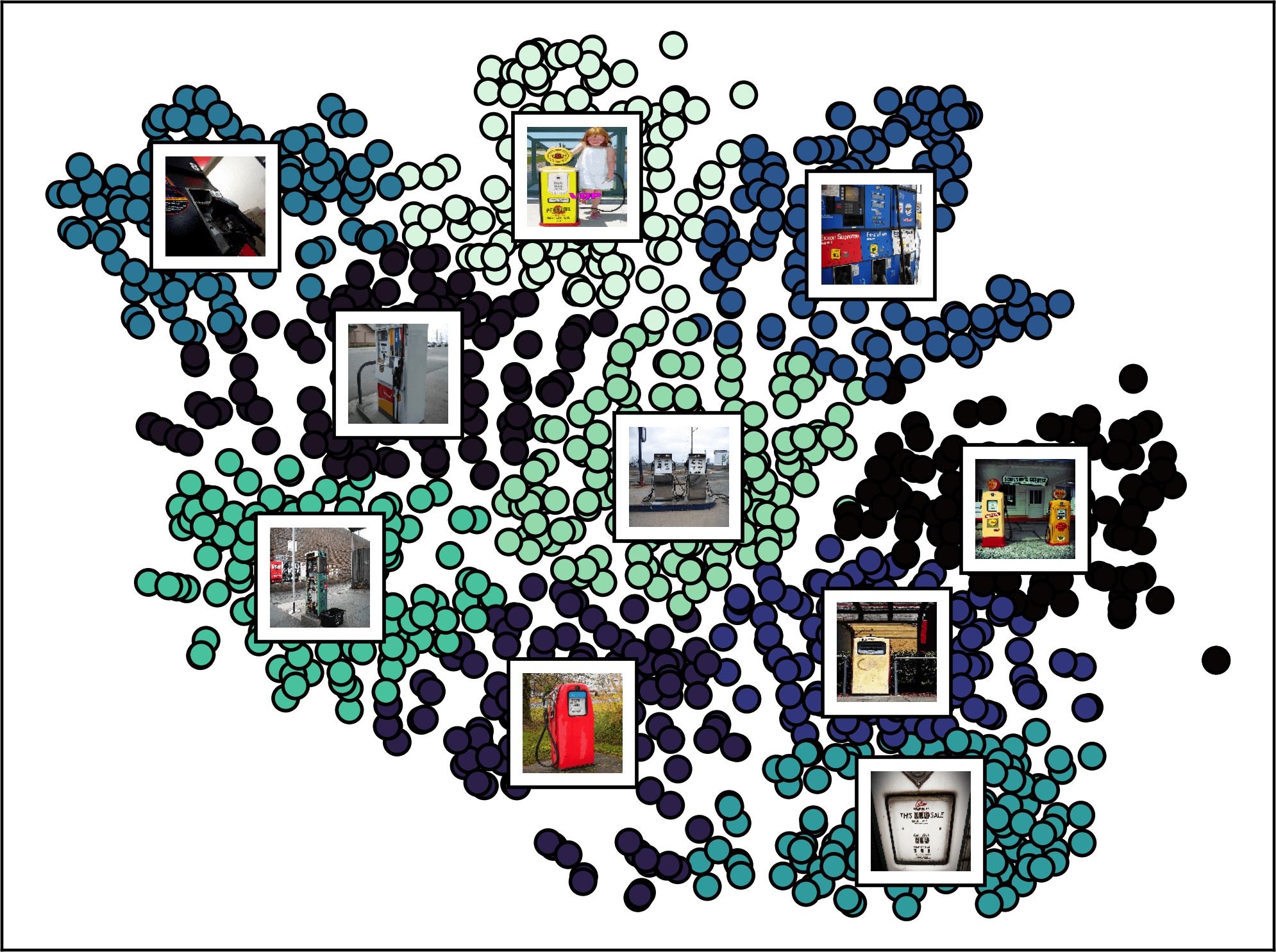}
        \includegraphics[trim=20px 20px 20px 20px, clip,width=\textwidth]{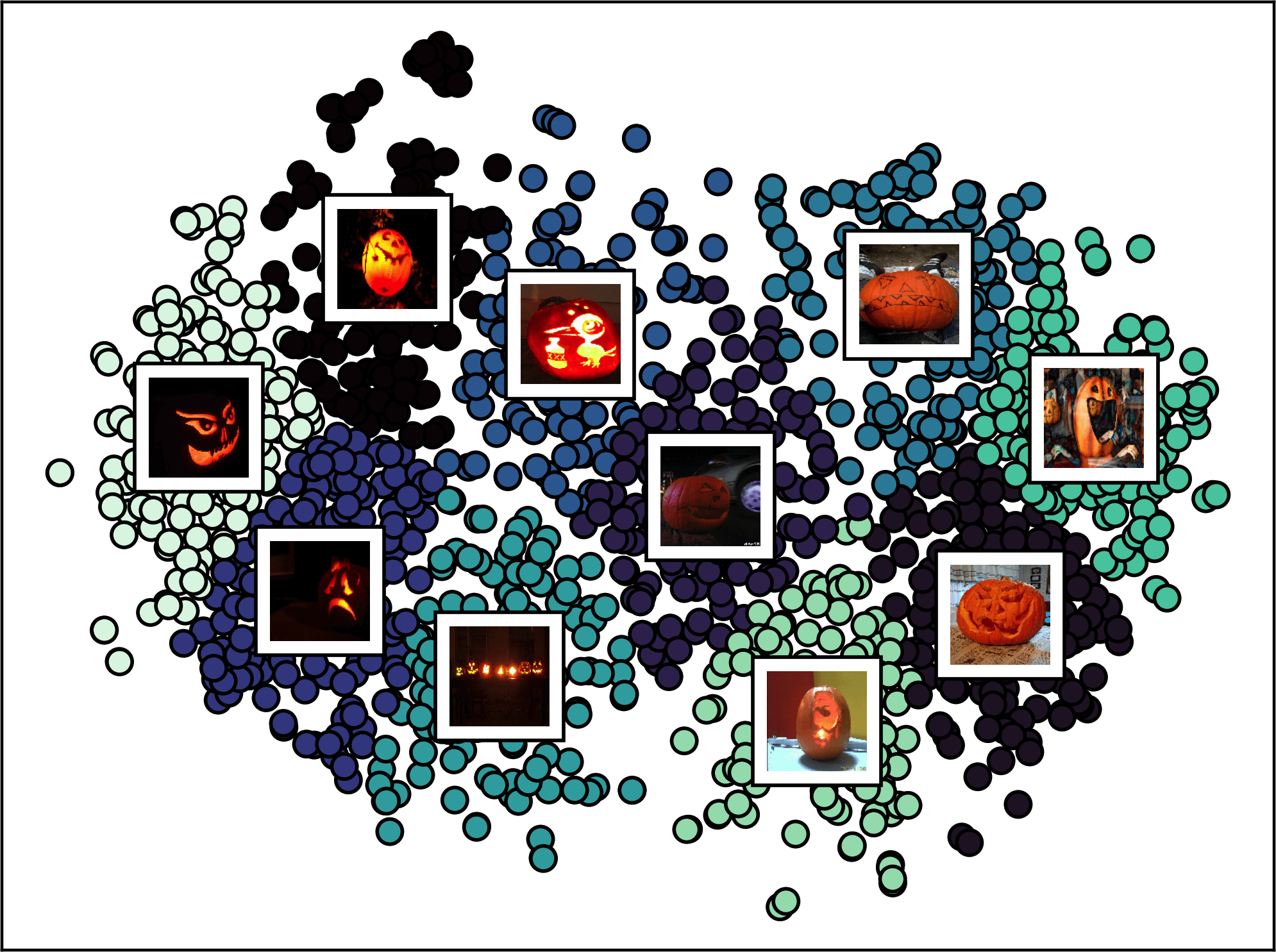}
        \captionsetup{labelformat=empty}
        \caption{a. Original}
        \label{fig:tsne_original_b}
    \end{minipage}
    \hfill
    \begin{minipage}[t]{.5\textwidth}
    \centering
        \includegraphics[trim=20px 20px 20px 20px, clip,width=\textwidth]{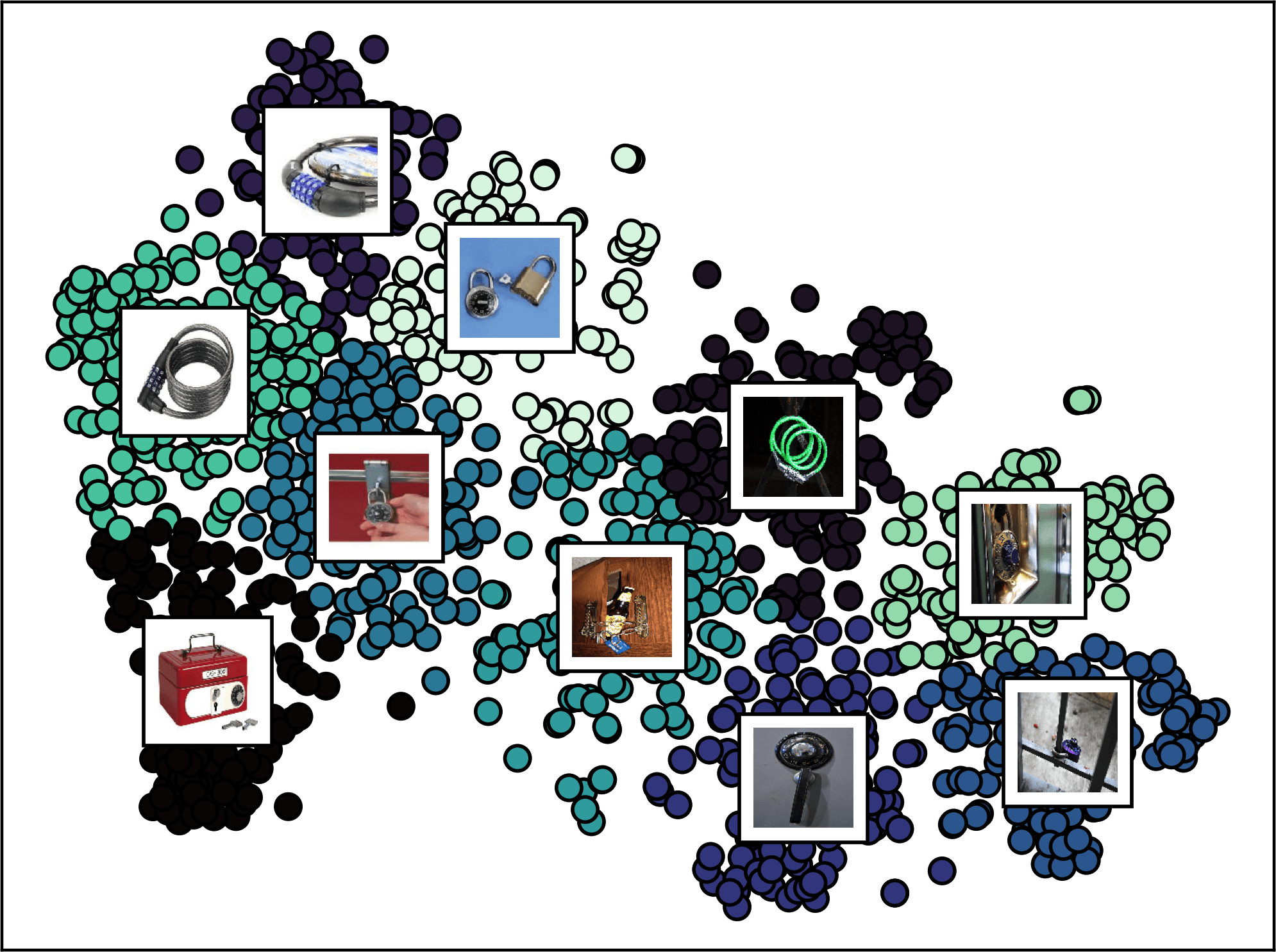}
        \includegraphics[trim=20px 20px 20px 20px, clip,width=\textwidth]{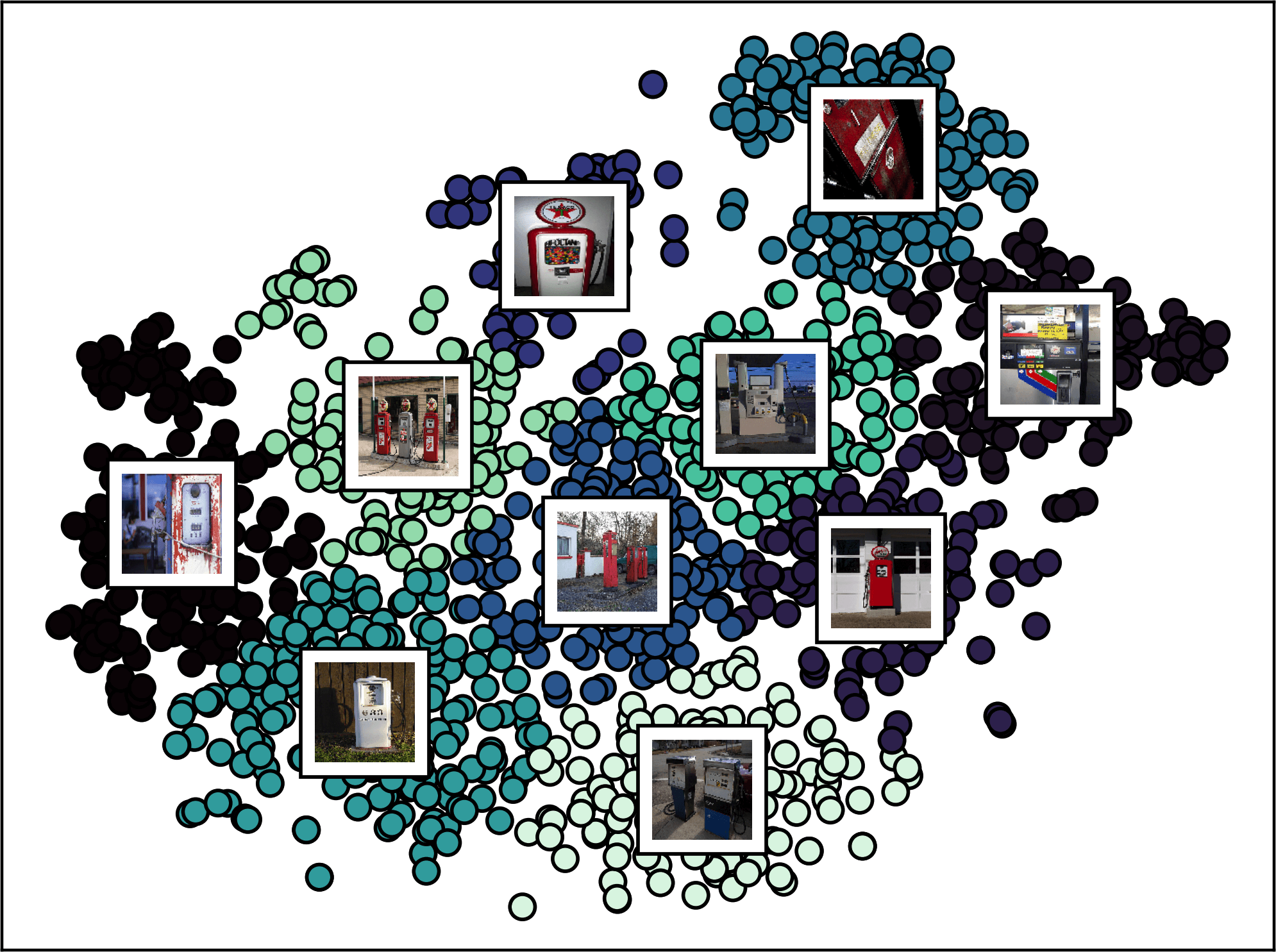}
        \includegraphics[trim=20px 20px 20px 20px, clip,width=\textwidth]{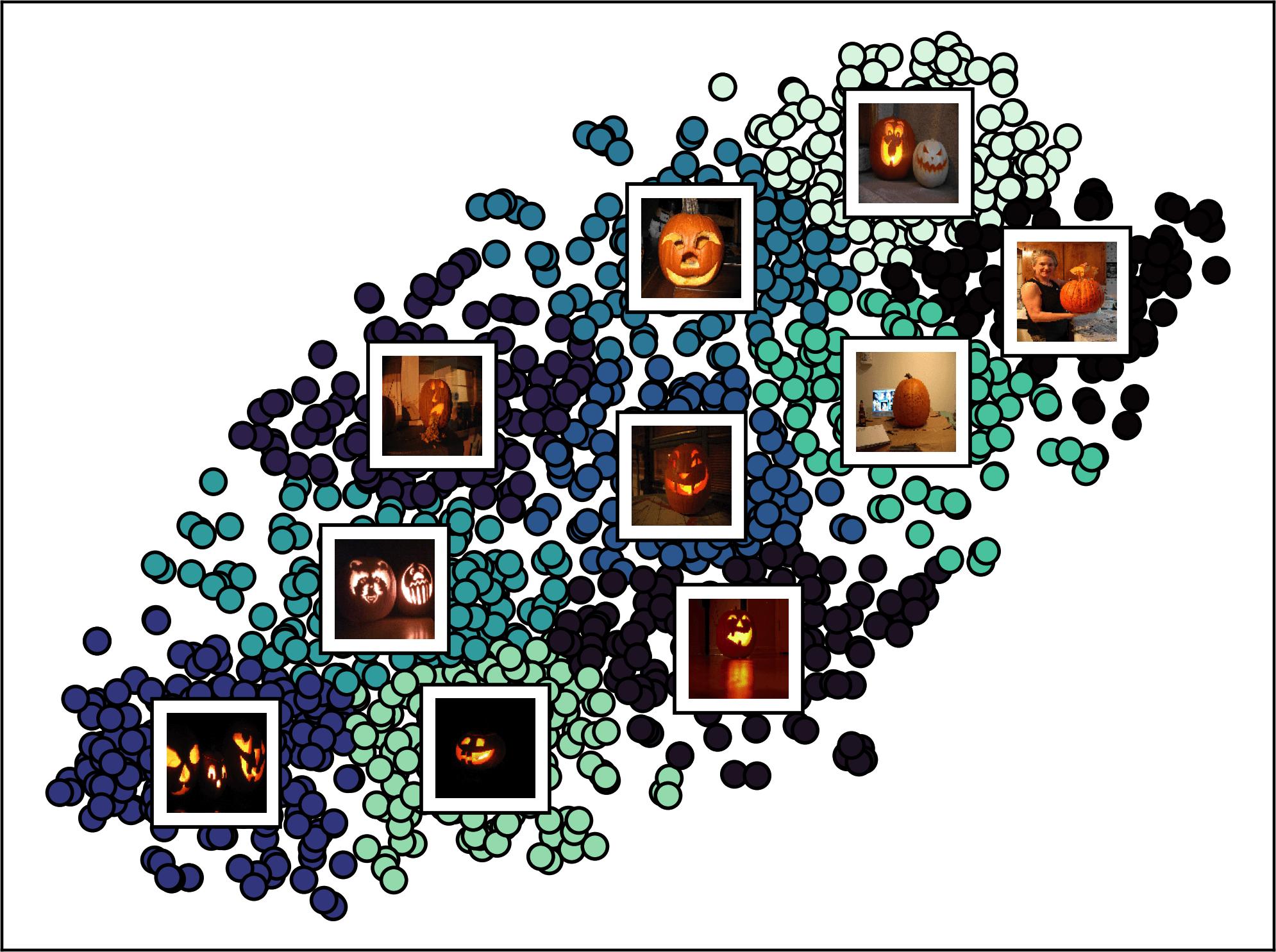}
        \captionsetup{labelformat=empty}
        \caption{b. SoftPool}
        \label{fig:tsne_softpool_b}
    \end{minipage}
\setcounter{figure}{5}
\caption{\textbf{t-SNE feature embeddings for InceptionV3 with and without SoftPool}. ImageNet1K classes ``combination lock'', ``gas pump'' and ``jack-o-lantern''. Cluster centers are found with k-means to better visualize the feature space. Images displayed are the closest examples for each cluster center.}
\label{fig:tsne_kmeans_b}
\end{figure*}

\begin{figure*}[!htb]
    \begin{minipage}[t]{.5\textwidth}
    \centering
        \includegraphics[trim=20px 20px 20px 20px, clip,width=\textwidth]{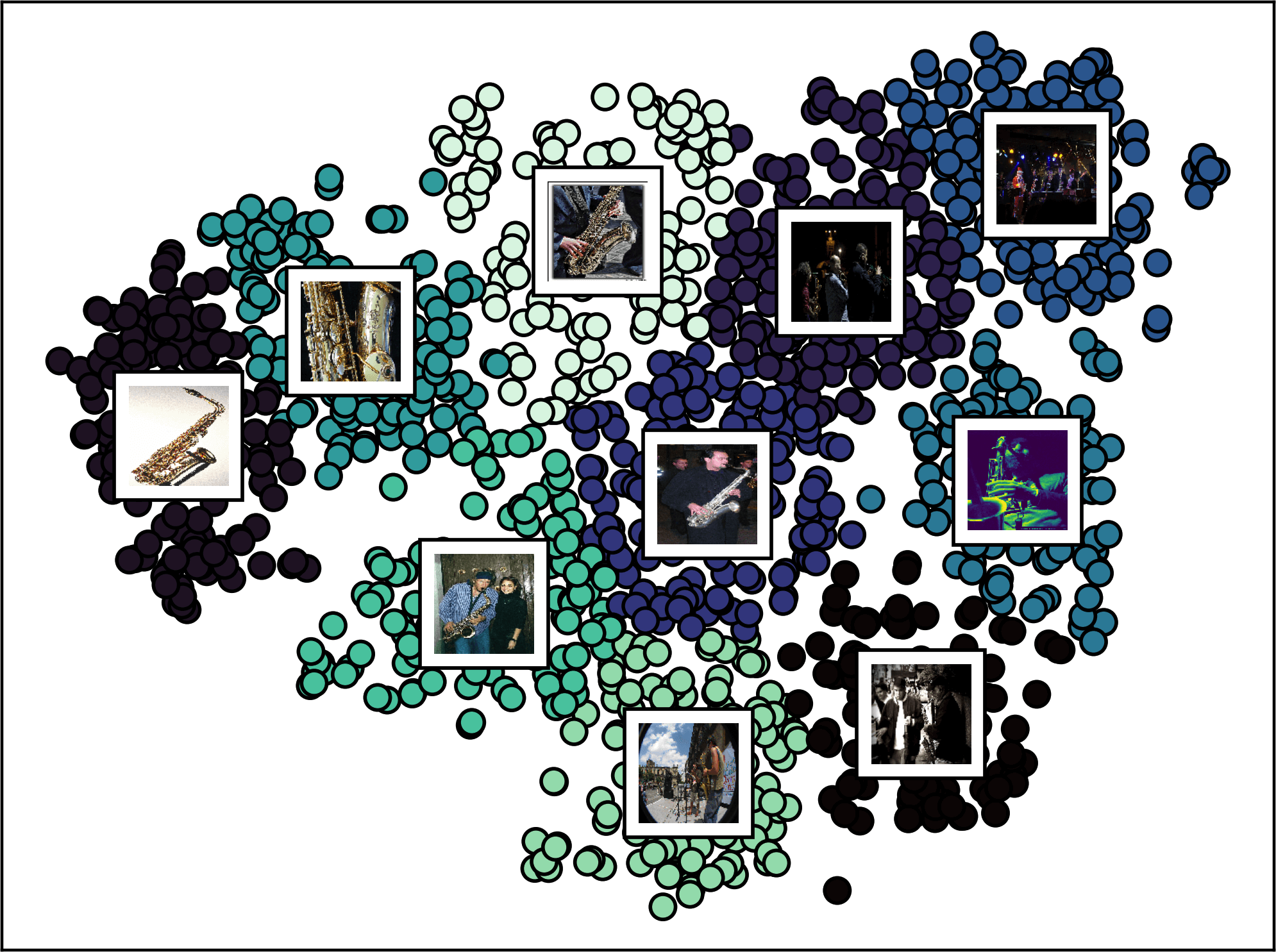}
        \includegraphics[trim=20px 20px 20px 20px, clip,width=\textwidth]{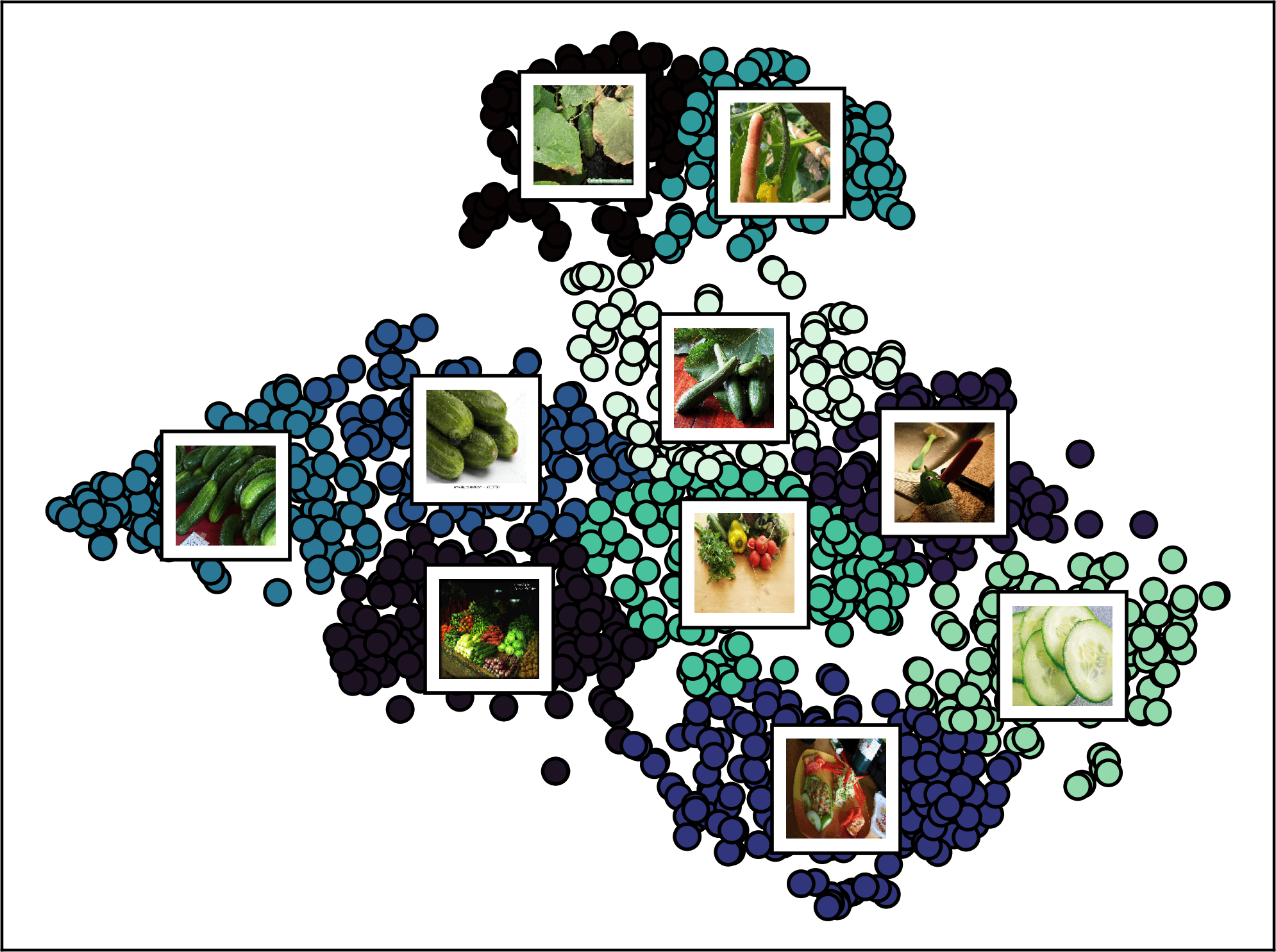}
        \captionsetup{labelformat=empty}
        \caption{a. Original}
        \label{fig:tsne_original_c}
    \end{minipage}
    \hfill
    \begin{minipage}[t]{.5\textwidth}
    \centering
        \includegraphics[trim=20px 20px 20px 20px, clip,width=\textwidth]{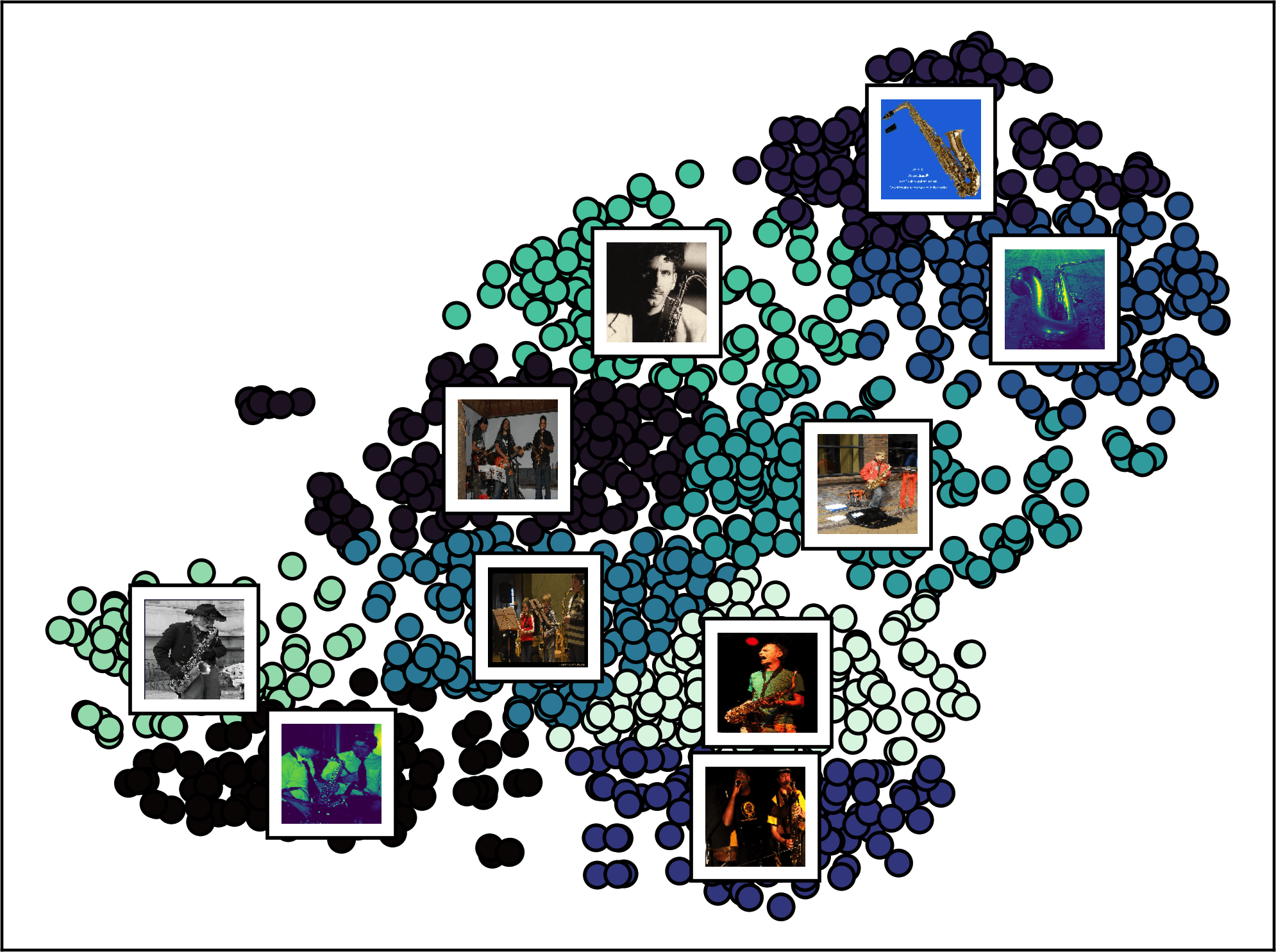}
        \includegraphics[trim=20px 20px 20px 20px, clip,width=\textwidth]{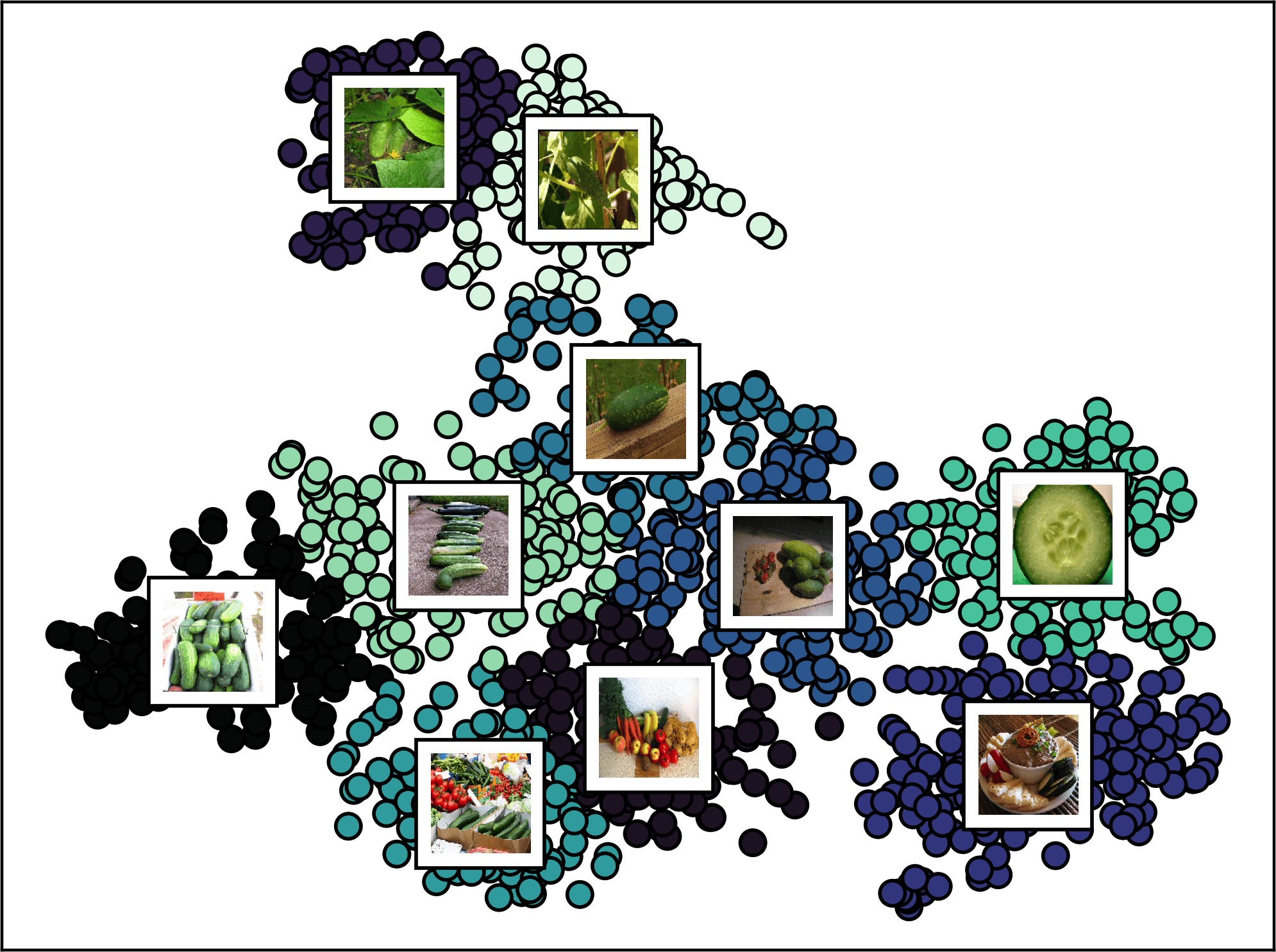}
        \captionsetup{labelformat=empty}
        \caption{b. SoftPool}
        \label{fig:tsne_softpool_c}
    \end{minipage}
\setcounter{figure}{6}
\caption{\textbf{t-SNE feature embeddings for InceptionV3 with and without SoftPool}. ImageNet1K classes ``sax'' and ``zucchini''. Cluster centers are found with k-means to better visualize the feature space. Images displayed are the closest examples for each cluster center.}
\label{fig:tsne_kmeans_c}
\end{figure*}

\end{document}